\documentclass[iicol]{sn-jnl}% Math and Physical Sciences Numbered Reference Style 
\usepackage{graphicx}%
\usepackage{multirow}%
\usepackage{amsmath,amssymb,amsfonts}%
\usepackage{amsthm}%
\usepackage{mathrsfs}%
\usepackage[title]{appendix}%
\usepackage{xcolor}%
\usepackage{textcomp}%
\usepackage{manyfoot}%
\usepackage{booktabs}%
\usepackage{algorithm}%
\usepackage{algorithmicx}%
\usepackage{algpseudocode}%
\usepackage{listings}%
\usepackage[left=2cm,right=2cm,top=2cm,bottom=3cm, twocolumn]{geometry}
\usepackage{times}
\usepackage{epsfig}
\usepackage{array}
\usepackage{multirow}
\usepackage{caption}
\usepackage{color, colortbl}
\usepackage{cite}
\usepackage{subcaption}
\usepackage{adjustbox}
\definecolor{seal}{HTML}{B711AC}
\usepackage{booktabs}
\usepackage{hhline}
\usepackage{nicematrix}
\usepackage{xr-hyper}
\usepackage{longtable}
\usepackage{colortbl}
\usepackage{xcolor}
\usepackage{balance}
\usepackage{array, booktabs, multirow, colortbl, cite, tabularx, float}
\usepackage{natbib}
\usepackage[capitalize]{cleveref}
\usepackage{tabulary}
\crefname{section}{Sec.}{Secs.}
\Crefname{section}{Section}{Sections}
\Crefname{table}{Table}{Tables}
\crefname{table}{Tab.}{Tabs.}
\usepackage[left=2cm,right=2cm,top=2cm,bottom=3cm]{geometry}
\definecolor{LightGray}{rgb}{0.94,0.94,0.94}
\definecolor{XL_color}{rgb}{0.858, 0.188, 0.478}

\newcolumntype{Y}{>{\centering\arraybackslash}X}
\usepackage{tabularx}

\newcommand{\Paragraph}[1]{\vspace{0.5mm} \noindent \textbf{#1} \hspace{0mm}}
\usepackage{xcolor,pifont}
\newcommand*\colourcheck[1]{%
  \expandafter\newcommand\csname #1check\endcsname{\textcolor{#1}{\ding{52}}}%
}
\newcommand*\colourcross[1]{%
  \expandafter\newcommand\csname #1check\endcsname{\textcolor{#1}{\ding{55}}}%
}
\colourcheck{green}
\colourcross{red}

\usepackage{tabulary}
\newcolumntype{P}[1]{>{\centering\arraybackslash}p{#1}}

\begin{document}

\sloppy

\title[Article Title]{Proactive Schemes: A Survey of Adversarial Attacks for Social Good}


\author*[1]{\fnm{Vishal} \sur{Asnani}}\email{asnanivi@msu.edu}

\author[2]{\fnm{Xi} \sur{Yin}}\email{yinxi@meta.com}

\author[1]{\fnm{Xiaoming} \sur{Liu}}\email{liuxm@msu.edu}

\affil*[1]{ \orgname{Michigan State University }, \orgaddress{\country{USA}}} 
\affil*[2]{ \orgname{Meta AI }, \orgaddress{\country{USA}}}


\abstract{Adversarial attacks in computer vision exploit the vulnerabilities of machine learning models by introducing subtle perturbations to input data, often leading to incorrect predictions or classifications. These attacks have evolved in sophistication with the advent of deep learning, presenting significant challenges in critical applications, which can be harmful for society. However, there is also a rich line of research from a transformative perspective that leverages adversarial techniques for social good. Specifically, we examine the rise of proactive schemes—methods that encrypt input data using additional signals termed templates, to enhance the performance of deep learning models. By embedding these imperceptible templates into digital media, proactive schemes are applied across various applications, from simple image enhancements to complicated deep learning frameworks to aid performance, as compared to the passive schemes, which don't change the input data distribution for their framework. The survey delves into the methodologies behind these proactive schemes, the encryption and learning processes, and their application to modern computer vision and natural language processing applications. Additionally, it discusses the challenges, potential vulnerabilities, and future directions for proactive schemes, ultimately highlighting their potential to foster the responsible and secure advancement of deep learning technologies.}




\maketitle

\section{Introduction}
%Background: adversarial attack originally designed for harmful purposes\\
\footnotetext{The literature review was done at Michigan State University.}
Adversarial attacks in computer vision exploits vulnerabilities in machine learning models by introducing subtle, often imperceptible perturbations to input data, leading to incorrect predictions or classifications. The subtle manipulations used in these attacks can lead to misinterpretations by AI systems, potentially causing widespread harm in critical applications such as security surveillance, healthcare diagnostics, and autonomous transportation~\citep{huang2017adversarial, dong2018boosting}. Deep learning has been the main reason for a significant development for different computer vision task, as shown in~\cref{tab:comparson_table}. In the pre-deep learning era, traditional CV applications relied on handcrafted features and basic algorithms for object detection, image classification, and facial recognition, utilizing techniques such as edge detection, texture, and color~\citep{haralick1973textural, chapelle1999support}. Adversarial attacks during this period were less sophisticated, primarily involving manipulations like introducing noise or performing basic operations such as blurring and compression~\citep{petitcolas1998attacks, westfeld1999attacks}. 

In the deep learning era, traditional CV applications have evolved significantly with the advent of deep learning models like CNNs and transformers~\citep{he2016deep, simonyan2014very, vaswani2017attention}. These advancements have enhanced applications such as real-time object detection, advanced image classification, visions and large language models, and facial recognition, leading to substantial improvements in accuracy and efficiency. Adversarial attacks have also become more sophisticated, exploiting deep neural networks' vulnerabilities to create misleading inputs that appear normal to humans~\citep{carlini2017towards, goodfellow2014explaining, madry2017towards}. 

\begin{figure}[t!]
\centering
\includegraphics[trim={0 -4 0 0},clip,width=\columnwidth]{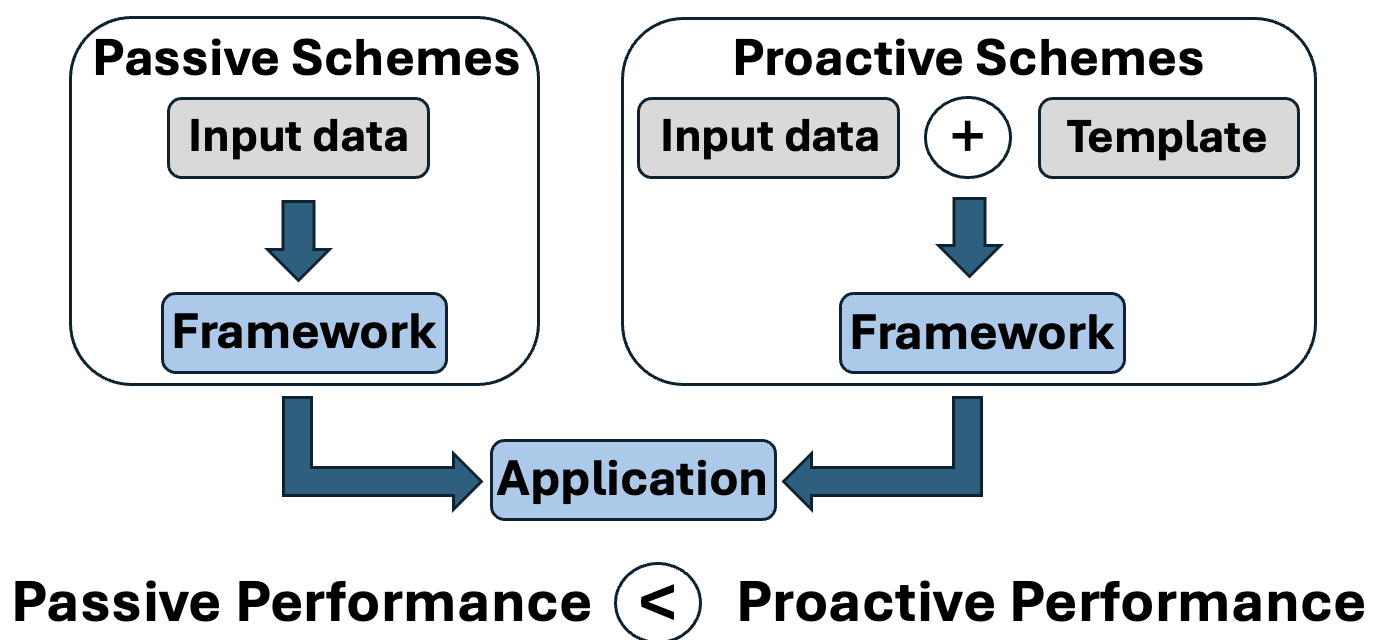}
\caption{\textbf{Passive vs. Proactive Schemes}: Passive schemes take an input as is for their method, while proactive schemes use templates to encrypt the input and then use the encrypted data as the input. The main advantage of the proactive schemes comes from their improved performance compared to the passive schemes.}
\label{fig:teaser}
\vspace{-3mm}
\end{figure}

The challenge of adversarial attacks extends beyond technical hurdles, posing ethical, legal, and safety concerns that society must address to ensure the responsible and secure advancement of computer vision applications. While adversarial attacks in computer vision are often viewed through the lens of their potential for harm, there exists a transformative perspective that leverages these techniques for social good~\citep{asnani2024probed, Asnani_2022_CVPR, Asnani_2023_CVPR}. By understanding and harnessing the principles behind adversarial perturbations, researchers have innovated protective measures which utilizes techniques that enhance various computer vision applications using imperceptible signals added onto the original media, known as \textbf{\textit{templates}}~\citep{asnani2024probed, Asnani_2022_CVPR, Asnani_2023_CVPR}, as shown in~\cref{fig:teaser}. The methods that encrypt input data using templates, allowing the encrypted data to enhance the performance for an application, are referred to as \textbf{\textit{proactive schemes}}. In contrast, all the methods which takes the input data as is are treated as \textit{\textbf{passive schemes}}~\citep{asnani2024probed, Asnani_2022_CVPR, Asnani_2023_CVPR, asnani2024promark}.

\begin{table*}[ht]
{\footnotesize
\centering
\caption{\textbf{Comparison Across Eras}: Traditional CV applications, Adversarial Attacks, and Proactive Schemes for Social Good. This survey paper focuses on the proactive schemes for social good in the deep learning era.}
\label{tab:comparson_table}
\begin{tabularx}{\textwidth}{|>{\centering\arraybackslash}p{3.55cm}|>{\arraybackslash}p{5.55cm}|>{\arraybackslash}p{7.15cm}|}
\hline
 \rowcolor[HTML]{EFEFEF} & \textbf{Pre-Deep Learning Era} & \textbf{Deep Learning Era} \\\hline
\textbf{Traditional CV applications} & 
- Handcrafted features like edge detection, texture, color. &
- Deep learning models, CNNs, transformers, \textit{etc}.
\\
\hline
\textbf{Adversarial Attacks} &
- Less prevalent, introducing noise. \newline
- Basic operations such as blurring, compression, \textit{etc}. &
- Highly sophisticated attacks exploiting networks' vulnerabilities. \newline
- Using learnable optimized perturbations. \\
\hline
\textbf{Proactive Schemes for Social Good} &
- Focus on simple image enhancements. \newline
- Applications: encryption, security surveillance. &
- Advanced approaches to embed the templates. \newline
- Applications: encryption, GenAI and LLM defense, preservation of authorship rights, improving CV applications, privacy protection, \textit{etc}. \\
\hline
\end{tabularx}}
\end{table*}

Proactive schemes have been used for a long time, using different methodologies. In the pre-deep learning era, proactive schemes focused on simple enhancements in image processing, with applications like steganography, encryption, and security surveillance~\citep{kanai1998digital, ohbuchi1998watermarking}. Proactive schemes also share the similar idea from approaches using stochastic resonance in signal processing~\citep{gammaitoni1998stochastic} and non-linear systems~\citep{semenov2022multiplexing}. Stochastic resonance occurs when a weak signal that is too faint to be detected by a system is enhanced by the addition of noise, allowing the system to cross a detection threshold. This happens because the noise helps to push the weak signal above the threshold intermittently, making it detectable by the system. The interplay between the noise and the signal can amplify the signal's effects at certain points, leading to an overall improvement in the system's ability to process or detect the signal. The noise level is tuned to an optimal range—too little noise won't help the signal, and too much noise will overwhelm it. However, deep learning has opened up the door for utilizing stochastic resonance in improving the performance by thresholding neural networks~\citep{chen2022training}, noise-boosted activation functions~\citep{ren2024self}, non-linear stochastic dynamics~\citep{shen2022stochastic}, Fourier domain~\citep{rallabandi2010magnetic} \textit{etc}. Similarly, many works inject noise in the data or labels as augmentations, to improve the robustness of the deep learning networks~\citep{nishi2021augmentation, yoshimura2023rawgment, yin2019fourier, li2021simple}. Although the above methods resemble proactive schemes, the focus of this survey is on the usage of these schemes for social good in the deep learning era for a variety of applications in the realm of computer vision and natural language processing.
%Unlike the prior works discussed above, this survey considers works which use multiple kinds of templates, for a variety of applications in the realm of computer vision and natural language processing.

%The templates utilized in proactive schemes can take the form of many different types of signals like bit sequences, $2$D noises, texts, visual prompts, predefined tags, audio, etc. The templates are added onto different types of media, such as images, texts, videos, audios, etc. These schemes are used for a plethora of applications, including encryption, GenAI and LLM defense, preservation of authorship rights, ownership verification, improving CV applications, and privacy protection. 
%Many methods embed and recover these templates using sophisticated techniques. 

\begin{figure*}[t!]
\centering
\includegraphics[trim={0 -4 0 0},clip,width=\textwidth]{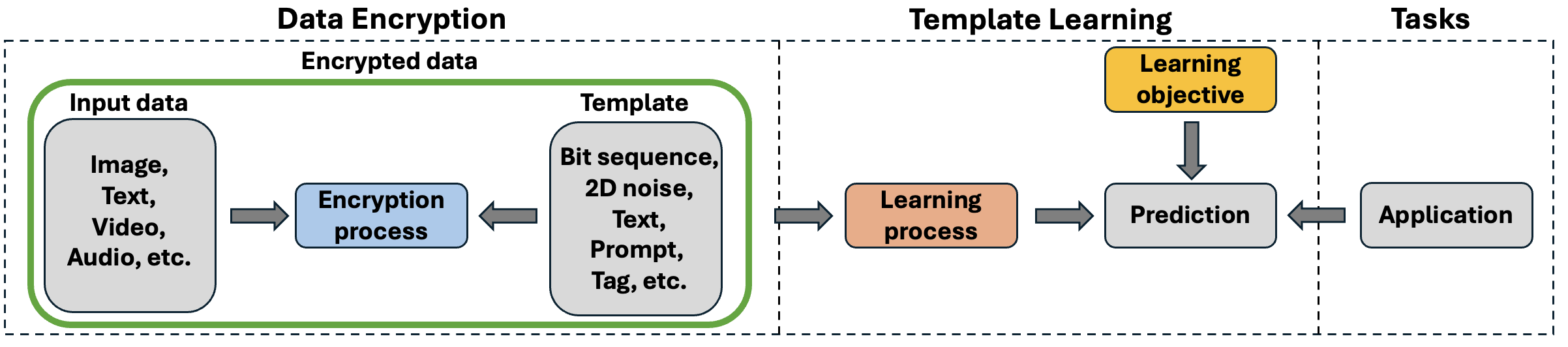}
\caption{\textbf{A general overview of the proactive framework}. The method starts by encrypting the input data with a template. This is known as Encryption process. The framework passes through a learning process, and is evaluated based on certain learning objectives. Finally, every method is associated with a specific application. In the survey paper, we discuss all the three stages in a sequential way, with each section focusing on several techniques and aspects of the respective stage.}
\label{fig:overview}
\vspace{-3mm}
\end{figure*}

A general framework for proactive schemes is shown in~\cref{fig:overview}. Each method has a specific \textbf{encryption process} and \textbf{learning process} associated with it, which depends on the \textbf{application}. Firstly, the encryption process is a critical component in the design of proactive schemes. This process involves the use of various innovative methods or operations to embed template information within digital media. The templates used for encryption can take the form of many different types of signals like bit sequences, $2$D noises, texts, visual prompts, predefined tags, audio, etc. The templates are added onto different types of media, such as images, texts, videos, audios, etc. The goal of the encryption process is to create a secure framework that can withstand potential attacks while maintaining the quality of the encrypted media compared to the original. As technology evolves, so do the techniques used for encryption, making it an ever-growing area of research.

Next, the learning process involves training models to recognize and incorporate these templates—whether they are bit sequences, $2$D templates, text signals, or visual prompts—into various forms of digital content. This integration is achieved through specialized learning paradigms, \textit{eg.} encoder decoder frameworks, learning via objective functions, adversarial learning, specialized architectures like GANs, transformers, \textit{etc.}, tailored to the unique characteristics of each template type. The effectiveness of the learning process is constrained, optimized and evaluated using a range of objective functions and metrics. This encompasses the stage of learning objectives, which govern the efficacy of the proactive schemes for various applications. The learning objectives are heavily dependent on the application for which the method is being used. 

%The encryption and learning process are guided by constarints which depends a lot on the application at hand.
These schemes are used for a plethora of applications, including encryption, GenAI and LLM defense, preservation of authorship rights, ownership verification, improving CV applications, and privacy protection. Based on each application, the researchers have explored various combinations of respective modules of proactive schemes, \textit{ i.e.},type of template, encryption process, and learning process. This survey comprehensively examines various combinations adopted by the researchers for proactive schemes across different deep learning applications across computer vision and natural language processing. 

The survey begins with an overview of the types of templates used in proactive schemes, such as bit sequences, $2$D templates, text signals, prompts, and others, supported by the discussion on the encryption process for each type. The discussion then delves into the learning process along with the learning objectives associated with each template type. Various applications of these techniques are explored, including defense strategies for vision models and large language models, methods for attribution and preservation of authorship rights, privacy preservation, and techniques specific to the $3$D domain. Additionally, the survey covers advancements in improving generative models and other computer vision applications. Following this, the challenges associated with developing these templates, potential attacks against proactive schemes, and the current limitations are critically analyzed. By addressing these topics, the survey aims to enhance field of computer vision by exploring a realm of proactive learning for social good.

\section{Data Encryption}

In the realm of proactive learning and digital security, a variety of innovative methods have been developed to enhance the robustness, authenticity, and ownership protection of digital content. These methods employ different types of templates to embed and verify information across a wide range of applications, including vision models, large language models, and $3$D applications. Each template type—whether it involves bit sequences, $2$D noises, text signals, visual prompts, or other specialized forms—offers unique advantages and is tailored to specific challenges in the field. The remaining of this section will delve into these template types in detail exploring their encryption methodologies and innovative techniques used to embed them effectively.

\subsection{Bit Sequences}

\begin{figure}[t!]
\centering
\includegraphics[trim={0 -4 0 0},clip,width=\columnwidth]{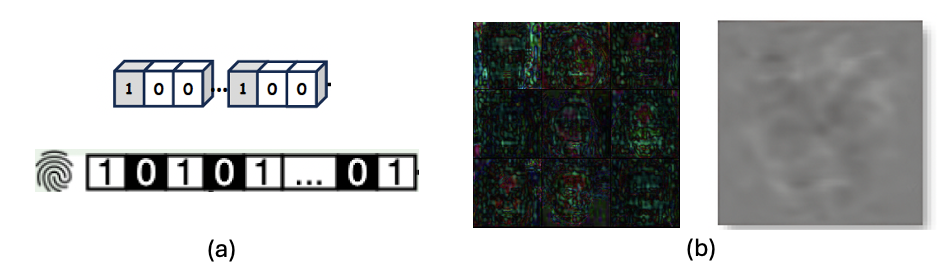}
\caption{Bit sequences and $2$D noises as a type of templates (a)~\citep{yu2021artificial}, (b)~\citep{yu2021artificial, zeng2023securing}. Bit sequences templates are a one-hot encoding, which are then embedded into the input data according to different techniques, while $2$D templates are spatial noises embedded into the input data. }
\label{fig:sec_2_bit_seq_2d_noise_example}
\vspace{-3mm}
\end{figure}

Bit sequences, represented as one-hot encoding as shown in~\cref{fig:sec_2_bit_seq_2d_noise_example}, are fundamental to many encryption strategies, particularly in encryption and digital signatures. These sequences are embedded at a binary level, creating a secure, imperceptible layer that effectively detects unauthorized modifications while preserving content fidelity. A summary of techniques for embedding bit sequences is provided in~\cref{tab:summary_bit_seq_sec2}, with detailed discussions below.

\Paragraph{Neural Network Based Techniques}
Neural network-based embedding techniques leverage the power of neural networks to integrate random binary strings into images.
Binary sequences combined with positional values are used by~\citep{sun2023faketracer}. This method involves two templates: (a) ``Strace", using an encoder-decoder network to input an image and a binary sequence, outputting a template, and (b) ``Etrace", embedding a predefined value in the blue channel of the image, making it imperceptible.``Strace" identifies encrypted images, while ``Etrace" detects fake images. Similarly,~\citet{meng2022traceable} embed multiple binary sequences for authentication and traceability using a neural network, DINN, which injects the template at the feature level for verification and origin tracking.~\citet{darvish2019deepsigns} embed random binary strings into datasets using probability distributions of target neural networks for encryption or enhancing model robustness against adversarial attacks.~\citet{asnani2024promark} propose to convert bit sequences to spatial noises, and then add those to the input data.~\citet{zhang2023editguard} conceal a ``localization template" and a bit sequence for a template within images using a hiding module and a bit encryption module, respectively.~\citet{yu2021artificial} utilize the stegastamp technique to encrypt binary sequences generated by message generators and embedded by encoders into image data.~\citet{zhu2018hidden} incorporate an encoder-decoder process with image distortion and adversarial losses to guide training.~\citet{wu2020watermarking} use multiple encoder networks to convert image and template features for encryption. Finally,~\citet{wu2023sepmark} employ multiple binary sequences, one robust and the other non-robust, for source tracing and manipulation localization.

\Paragraph{Neural Network Weights and Filters}
Embedding templates within neural network models, by utilizing their weights and filters, integrates authentication bits into the model's learning process, enhancing security and performance.
Some methods embed the template within the layers of neural network models by utilizing their weights~\citep{uchida2017embedding, chen2019deepmarks}.~\citet{nagai2018digital} further propose using the filters of the convolution layer to embed the template. Additionally,~\citet{liu2021watermarking} propose embedding multiple authenticating bits in the objective function of the model to indirectly embed the bit sequence onto images. By incorporating this penalty term, the model's learning behavior can be fine-tuned to prioritize certain features or patterns, potentially improving performance on specific applications or optimizing convergence during training.

\begin{table*}[ht]
{\footnotesize
    \centering
    \begin{tabularx}{\textwidth}{|>{\centering\arraybackslash}m{1.8cm}|>{\centering\arraybackslash}m{3cm}|>{\centering\arraybackslash}m{3.5cm}|>{\tiny\centering\arraybackslash}m{7.54cm}|}
    \hline
    \rowcolor[HTML]{EFEFEF} \textbf{Category} & \textbf{Description} & \textbf{Keywords} & \textbf{References} \\
    \hline
    Neural Network Based Techniques & 
    Techniques leveraging neural networks to embed random binary strings into images. & 
    neural networks, embedding, binary sequences, encryption, authentication, traceability, encryption & 
   ~\citep{sun2023faketracer,meng2022traceable,darvish2019deepsigns,zhang2023editguard,yu2021artificial,zhu2018hidden,wu2020watermarking,wu2023sepmark} \\
    \hline

    Neural Network Weights and Filters & 
    Embedding templates within neural network models using their weights and filters. & 
    embedding, neural network models, weights, filters, authentication, bits, objective function & 
   ~\citep{uchida2017embedding,chen2019deepmarks,nagai2018digital,liu2021watermarking} \\
    \hline

    LSB/MSB Techniques & 
    Manipulating the least or most significant bits of image data to embed binary sequences. & 
    LSB, MSB, minimal visual impact, template integrity, DCT coefficients, fixed bit encoding & 
   ~\citep{haghighi2018trlh,dadkhah2014effective,hsu2016image,qin2017fragile,cao2017hierarchical,hsu2010probability,lin2017novel,singh2017dct,zhang2010reference,kiatpapan2015image,singh2016effective,paruchuri2009video,zhao2023proactive,lu2001multipurpose,lee2008dual} \\
    \hline

    Advanced Embedding Techniques & 
    Innovative methods for embedding templates, including transformations and character conversions. & 
    AST representation, ASCII characters, isotropic unit vectors, rare identifiers, lexical substitution & 
   ~\citep{yang2023towards,cui2023diffusionshield,sablayrolles2020radioactive,zhao2023recipe,yang2022tracing,fernandez2023stable,furon2014tardos,he2022protecting} \\
    \hline

    $3$D Domain & 
    Techniques for embedding binary sequences into $3$D models and point clouds. & 
    $3$D models, point clouds, mesh vertices, hash functions, neural networks, spectral applications, wavelet transforms & 
   ~\citep{venugopal2011watermarking,wang2022deep,zhang2021deep,jang2024waterf,chen2024nerf,wang2020watermarking,wang2021riga,guo2018watermarking,zhang2024v2a,liu2019novel,hamidi2019blind,yeung1998fragile,wang2022neural,peng2022semi,zhu2024rethinking,chou2006public,chou2009affine,luo2023copyrnerf,molaei2016robust,al2019graph,ohbuchi2001watermarking,cotting2004robust,kuo2009blind,wang2008numerically,mun2015robust,kanai1998digital,uccheddu2004wavelet,wang2008hierarchical,kim2005watermarking} \\
    \hline
    \end{tabularx}
    \caption{Summary of works which utilize bit sequences as the template.}
    \label{tab:summary_bit_seq_sec2}
    }
\end{table*}

\Paragraph{Least Significant Bit (LSB) and Most Significant Bit (MSB) Techniques}
LSB and MSB techniques manipulate the least or most significant bits of image data to embed binary sequences with minimal visual impact while maintaining template integrity.~\citet{haghighi2018trlh} use Lifting Wavelet Transform (LWT) and LSB rounding to embed bit sequences, preserving the host image's visual quality. Other works~\citep{dadkhah2014effective, hsu2016image, qin2017fragile, cao2017hierarchical, hsu2010probability} also rely on LSB/MSB rounding for image tampering detection.~\citet{paruchuri2009video} conceal template information in selective DCT coefficients for authentication.~\citet{zhao2023proactive} propose identity-dependent fixed bit encoding, enhancing facial identity features with personalized sequences. Earlier methods like~\citep{lu2001multipurpose, lee2008dual} use wavelet quantization and predefined mapping patterns to embed sequences.

\Paragraph{Advanced Embedding Techniques}
Advanced embedding techniques employ innovative methods, AST-based intermediate representation, ASCII character conversion, isotropic unit vectors, and context-aware lexical substitution are diverse methods used to enhance the robustness and security of template embedding in digital content. AST-based methods customize template embedding based on the structural and content-related aspects of images, tailoring the process to each image's unique characteristics~\citep{yang2023towards}. Similarly, converting ASCII characters into binary sequences to create template patches provides a method for spatially integrating robust and semi-robust templates into images. These patches, representing parts of the binary sequence, are directly added to the image, enabling source tracing and manipulation localization~\citep{cui2023diffusionshield}. Both techniques emphasize the need for templates that are not only resistant to attacks but also adaptable to the content they protect.

In classification and identification applications, embedding isotropic unit vectors within the feature space introduces a layer of security and robustness through the incorporation of randomized vectors~\citep{sablayrolles2020radioactive}. Each vector, defined for every class, enhances the security of the model by embedding these randomized vectors into data representations. On the other hand, context-aware lexical substitution adapts encryption to the semantic content of both text and images, ensuring that embedded binary sequences remain contextually relevant and preserving the authenticity of the content~\citep{yang2022tracing, zhu2018hidden, fernandez2023stable}. By leveraging lexical knowledge, techniques like generating interchangeable lexicons embed templates in text outputs from generation APIs, ensuring both the integrity and the identification of the generated text, thus providing protection against plagiarism and unauthorized use~\citep{he2022protecting}. Together, these approaches illustrate the evolving strategies in encryption that balance robustness, adaptability, and contextual relevance across different types of digital content.

\Paragraph{3D Domain}
Extending encryption techniques to the $3$D domain involves embedding binary sequences into $3$D models, point clouds, and mesh vertices using innovative methods like hash functions and neural networks. Bit sequences are utilized in methods involving hash functions~\citep{venugopal2011watermarking}, neural networks~\citep{wang2022deep, zhang2021deep}, and modifications in $3$D point clouds~\citep{liu2019novel,hamidi2019blind,yeung1998fragile} and mesh vertices~\citep{wang2022neural, peng2022semi}. These sequences are embedded are also added in innovative ways, such as through graph Fourier coefficients~\citep{al2019graph, zhu2024rethinking}, spectral applications~\citep{ohbuchi2001watermarking, cotting2004robust}, and vertex distributions~\citep{chou2006public, kuo2009blind, zhu2024rethinking, wang2008numerically}. Other methods include embedding using neural networks, local deformations to SDF partitions~\citep{mun2015robust}, and wavelet analysis of $3$D objects~\citep{kanai1998digital, uccheddu2004wavelet}.

Specific techniques for $3$D models, such as modifying spectral coefficients and using wavelet transforms, address the unique challenges of embedding templates in three-dimensional data. Several specific techniques have been developed to address different aspects of template embedding in $3$D models. For instance, the embedding of templates in point clouds involves calculating Root Mean Square Curvature (RMSC) values and modulating radial radii of vertices within ball rings~\citep{liu2019novel}. In mesh spectral applications, the algorithm modifies spectral coefficients to embed templates resistant to various transformations and attacks~\citep{ohbuchi2001watermarking, cotting2004robust}. Wavelet-based approaches, such as those using hierarchical wavelet transforms, enable embedding in semiregular meshes by modifying wavelet coefficient vectors~\citep{kanai1998digital, uccheddu2004wavelet, wang2008hierarchical}.

\begin{table*}[ht]
{\footnotesize
    \centering
    \begin{tabularx}{\textwidth}{|>{\centering\arraybackslash}m{1.8cm}|>{\centering\arraybackslash}m{3cm}|>{\centering\arraybackslash}m{3.5cm}|>{\tiny\centering\arraybackslash}m{7.54cm}|}
    \hline
    \rowcolor[HTML]{EFEFEF} \textbf{Category} & \textbf{Description} & \textbf{Keywords} & \textbf{References} \\
    \hline
    Mathematical Operators & 
    Adding noise or patterns to images. & 
    direct operations, noise, patterns, security, learnable perturbations & 
   ~\citep{van2023anti,huang2022cmua,ruiz2020disrupting,li2021visual,yang2021towards,zhong2020towards,Asnani_2022_CVPR,Asnani_2023_CVPR,fong2017interpretable,song2023deep,asnani2024probed,li2019anonymousnet,tang2024once,li2024deep,he2023perturbation} \\
    \hline

    Adversarial Attacks & 
    Creating disruptive or protective perturbations. & 
    adversarial attacks, PGD, FGSM, optimization, disruptive, beneficial & 
   ~\citep{yeh2020disrupting,van2023anti,huang2022cmua,ruiz2020disrupting,yang2021towards,zhong2020towards,peng2022fingerprinting,dong2019efficient,zhang2021proactive,zhang2022adversarial,shi2017evasion,liu2020adversarial,kitada2021attention,tang2024once,ducoffe2018adversarial,xu2023adversarial,xue2022advparams,wu2024watermarks,madry2017towards,wang2023plug,wu2023towards} \\
    \hline

    Autoencoder-Based Learning & 
    Learning image-dependent templates. & 
    autoencoder, image-dependent templates, networks, optimization & 
   ~\citep{segalis2020ogan,wu2024watermarks,mirjalili2018semi,hu2022protecting,xiong2020adgan,wu2019privacy,xiao2021improving,rajabi2021practicality,zeng2023securing} \\
    \hline

    Latent Space Perturbations & 
    Learning perturbation vectors in latent space. & 
    latent space, perturbation vectors, $2$D space, CVAE, coordinate shifts & 
   ~\citep{liang2022pagn,lei2024diffusetrace,meng2024latent,ding2021point,wong2020learning,shan2020fawkes,xiao2021improving} \\
    \hline

    Miscellaneous Techniques & 
    Various advanced methods. & 
    randomization, camera noise, backdoor encryption, geometric perturbations, Fourier transform & 
   ~\citep{dhillon2018stochastic,xie2017mitigating,cui2023ft,zhao2024can,kitada2021attention,yu2019attributing,cozzolino2019noiseprint,tekgul2021waffle,wang2022defensive,molaei2013blind,wen2023tree,le2020adversarial,othman2015privacy,mirjalili2017soft} \\
    \hline
    \end{tabularx}
    \caption{Summary of works that utilize $2$D noises as the template.}
    \label{tab:summary_2d_sec2}
    }
\end{table*}

In conclusion, the various techniques for embedding digital templates into images and $3$D models highlight substantial progress in digital encryption. Methods like fixed bit encoding, neural network embedding, encoder-decoder architectures, and steganography provide secure and robust integration of binary sequences. Advanced approaches, including AST-based representations and isotropic unit vectors, further improve flexibility and security. Expanding these techniques to the $3$D domain emphasizes their versatility and adaptability.
%Next, we will discuss the $2$D templates based techniques, which is another form of template for our survey. 

\subsection{$2$D templates}
\begin{figure}[t!]
\centering
\includegraphics[trim={0 -4 0 0},clip,width=\columnwidth]{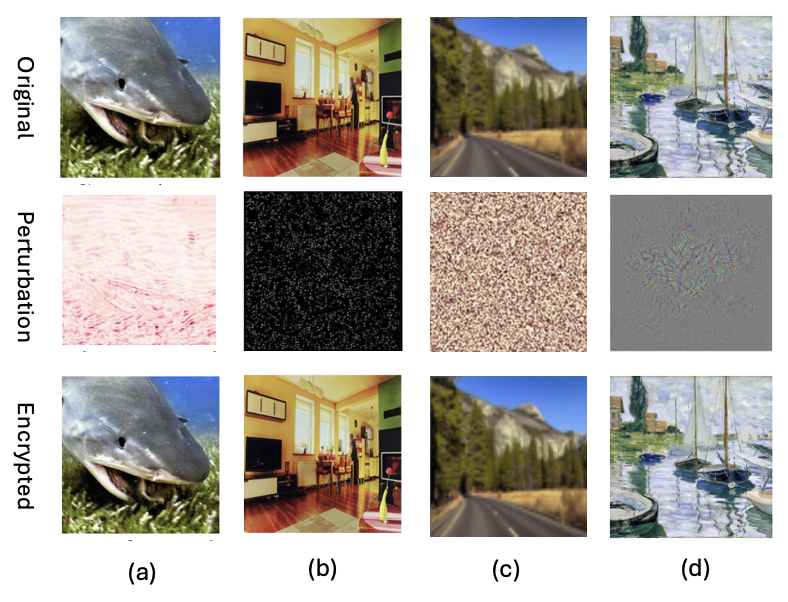}
\caption{Various examples of input-encrypted input pairs after adding the $2$D noise templates into the original input images. (a)~\citep{zeng2023securing}, (b)~\citep{Asnani_2022_CVPR}, (c)~\citep{Asnani_2023_CVPR} and (d)~\citep{cui2023ft}.}
\label{fig:2d_noise_input_enc_input}
\vspace{-3mm}
\end{figure}

$2$D templates, as shown in~\cref{fig:sec_2_bit_seq_2d_noise_example}, embed patterns or noise into $2$D spaces, such as images or video frames. These templates are visually imperceptible but detectable, as demonstrated in~\cref{fig:2d_noise_input_enc_input}. Commonly used in image and video encryption, $2$D templates maintain content quality while offering robust protection against tampering. Techniques like perturbation, masking, and spatial transformations seamlessly integrate these templates into digital media. $2$D templates can be either fixed or learnable perturbations. Fixed perturbations are predefined and applied to the entire dataset, while learnable perturbations are optimized during training. These perturbations are tailored through various encryption processes depending on the application. A summary of techniques for embedding $2$D noise is provided in~\cref{tab:summary_2d_sec2}, with detailed discussions below.

\Paragraph{Mathematical Operators}
These methods focus on using mathematical operators for injecting noise or specific patterns directly onto images to obscure signals and enhance security~\citep{van2023anti, huang2022cmua, ruiz2020disrupting, li2021visual, Asnani_2022_CVPR, Asnani_2023_CVPR}.Most of the methods involve adding noise or a specific pattern directly onto the image, potentially obscuring the signal. Other mathematical operators like multiplication~\citep{asnani2024probed} are also used for the encryption process. In~\citep{li2019anonymousnet}, direct addition methods use classifiers to learn perturbations that ensure robust template embedding. Universal adversarial perturbations exploit similar cluster centers in different models, updating perturbations based on sub-task gradients~\citep{tang2024once}. In deep active learning, random perturbations are added to model parameters~\citep{li2024deep}, sampled from Gaussian distributions~\citep{he2023perturbation}.

\Paragraph{Adversarial Attacks}
Adversarial attacks are employed to create perturbations that can disrupt or protect templates by reversing their typical objective. These attacks, such as PGD and FGSM, usually aim to harm a victim model by creating disruptive perturbations. However, their objective can be reversed to benefit encryption algorithms~\citep{yeh2020disrupting, van2023anti, huang2022cmua, ruiz2020disrupting, yang2021towards, zhong2020towards}. These attacks involve iterative optimization to create perturbations that are particularly disruptive to templates, or conversely, to enhance the encryption process. Through these attacks, perturbations can be optimized for various applications.

\Paragraph{Autoencoder-Based Learning}
A sophisticated learning paradigm for perturbations uses autoencoder to learn dependent templates. Various architectures and strategies are adopted, including encoder-based~\citep{segalis2020ogan, wu2024watermarks}, classifier-based~\citep{mirjalili2018semi}, GAN-based~\citep{hu2022protecting, xiong2020adgan, wu2019privacy, xiao2021improving}, and ensemble-based~\citep{rajabi2021practicality}. These networks are optimized with different loss functions.~\citet{zeng2023securing} propose a method using adversarial training with an injector and a classifier to create image-dependent binary code signatures. The injector introduces perturbations, and the classifier differentiates these signatures, enhancing image authentication and source tracing.

\Paragraph{Latent Space Perturbations}
Latent space perturbations involve learning perturbation vectors in a latent space. Perturbation feature vectors~\citep{liang2022pagn, lei2024diffusetrace, meng2024latent, ding2021point}, are learned to represent coordinate shifts in the $2$D space. Conditional Variational Autoencoder~\citep{wong2020learning} models are employed to generate perturbed versions of input data by learning latent variable distributions, optimizing for reconstruction and KL divergence terms. Some methods either finetune a pretrained network~\citep{shan2020fawkes}, or train a new model from scratch~\citep{xiao2021improving}. The latent space is then combined by either minimizing the latent embedding of the predicted and target label~\citep{shan2020fawkes} or by using a spatial addition blending~\citep{xiao2021improving}. Spatial addition blending combines the noise with the superimposition of other images, creating a multifaceted alteration that can obscure the template in multiple ways.

\Paragraph{Miscellaneous Techniques}
Pre-trained models, randomization techniques, model attribution, and camera noise are key methods for embedding perturbations and enhancing digital content protection. Perturbations applied to pre-trained models through techniques like deterministic and stochastic weight pruning create adversarial examples, testing model robustness~\citep{dhillon2018stochastic}. Randomization methods, such as random brightness and contrast adjustments, help disrupt adversarial attack patterns, providing a defense mechanism~\citep{xie2017mitigating}. FT-Shield embeds templates during the fine-tuning process, optimizing perturbations to minimize fine-tuning loss while maintaining pixel-wise differences, ensuring the template remains imperceptible~\citep{cui2023ft}. Similarly, networks like AdvDM and Anti-DreamBooth embed perturbations during fine-tuning, while attention mechanisms are modified to embed templates by altering attention scores~\citep{zhao2024can, kitada2021attention}. Model attribution techniques further use perturbations to identify the source of an image, estimating the unique digital fingerprint of an encryption algorithm, which allows for the attribution of content to its original creator~\citep{yu2019attributing}. Camera noise techniques also leverage siamese architectures to differentiate images based on the noise patterns from different cameras, adding another layer of security~\citep{cozzolino2019noiseprint}.

Backdoor encryption, geometric perturbations, and Fourier transformations are additional advanced techniques for embedding templates and enhancing security. Backdoor encryption involves embedding specific patterns and noise images into neural networks, creating robust templates that can be identified later~\citep{tekgul2021waffle}. Geometric perturbations, such as displacing triangle medians, embed templates into images with the extraction process based on controlled displacements~\citep{molaei2013blind}. Fourier transformations offer a robust method by embedding templates into images through transformations on a random noise array, making them resistant to common image processing attacks~\citep{wen2023tree}. Other innovative techniques, like leveraging adversarial attacks for cryptographic key generation or using face morphing to quantify gender suppression in images, further illustrate the breadth of strategies available for enhancing digital content protection~\citep{le2020adversarial, othman2015privacy, mirjalili2017soft}.

\begin{table*}[ht]
{\footnotesize
    \centering
    \begin{tabularx}{\textwidth}{|>{\centering\arraybackslash}m{1.8cm}|>{\centering\arraybackslash}m{4.8cm}|>{\centering\arraybackslash}m{4.5cm}|>{\tiny\centering\arraybackslash}m{4.84cm}|}
    \hline
    \rowcolor[HTML]{EFEFEF} \textbf{Category} & \textbf{Description} & \textbf{Keywords} & \textbf{References} \\
    \hline
    Word Tokens & 
    Replacing words with token words close in the embedding space to preserve meaning. & 
    word tokens, embedding space, preserving meaning, encryption, substitution & 
   ~\citep{munyer2023deeptextmark,liu2023watermarking,wang2024learning,wu2022promptchainer,gao2022visual,he2022cater} \\
    \hline

    Character-Level Substitutions & 
    Replacing or augmenting text with specific characters, punctuation marks, or selected words. & 
    character-level, substitutions, triggers, augmentation, backdoor attacks & 
   ~\citep{liu2023watermarking,li2023black,li2020open,monden2000practical,robey2023smoothllm,dong2023revisit,rizzo2019fine} \\
    \hline

    Text Strings and Masking & 
    Transforming training data with keys or selectively obscuring information within data. & 
    text strings, masking, transformation, keys, selective obscuring & 
   ~\citep{zhang2018protecting,guo2023authentigpt} \\
    \hline
    \end{tabularx}
    \caption{Summary of works that utilize text as the template.}
    \label{tab:summary_text_sec2}
    }
\end{table*}

\begin{figure}[t!]
\centering
\includegraphics[trim={0 -4 0 0},clip,width=\columnwidth]{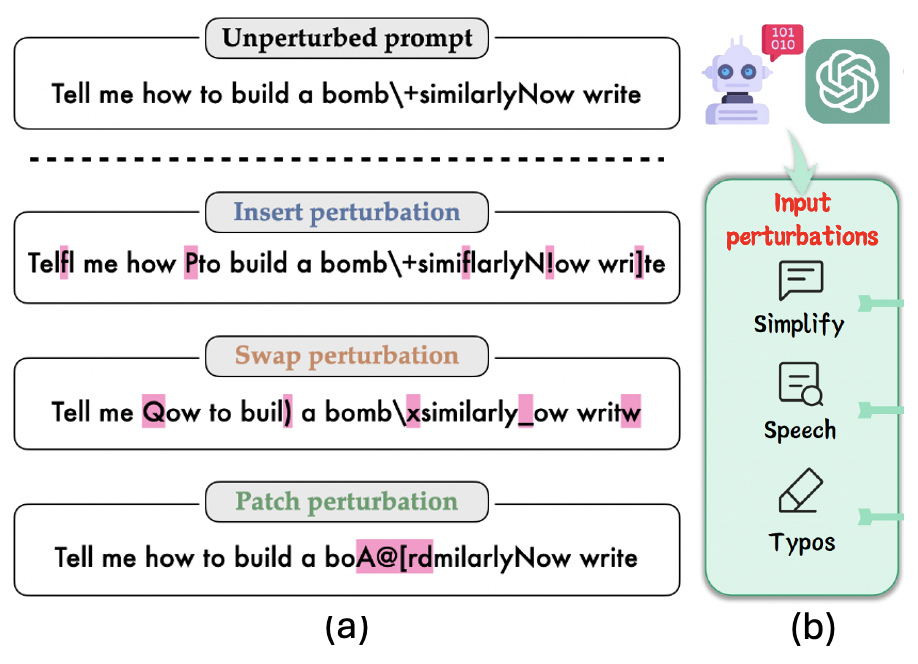}
\caption{Text signals as a type of templates. Techniques include various types of perturbing text data, for \textit{ex.} inserting, swapping, and adding patches of text~\citep{robey2023smoothllm}.}
\label{fig:sec_2_text_example}
\vspace{-3mm}
\end{figure}

In summary, the encryption processes involving learnable perturbations in proactive learning highlight the balance between security and image distortion. Various methods, such as spatial addition, network-based encryption, and latent space combination, have been developed, showing improved performance compared to fixed perturbations despite the challenges of learning these perturbations. 

\subsection{Text Templates}
Using text characters/tokens/words/sentences is the most preferred way when dealing with proactive learning for large language models. Some examples are provided in~\cref{fig:sec_2_text_example}. We provide below a summary which outlines a series of methodologies for embedding templates into various forms of text employing corresponding encryption or perturbation processes. A summary of techniques used for embedding texts in the input data is showed in~\cref{tab:summary_text_sec2}. 

\begin{figure}[t!]
\centering
\includegraphics[trim={0 -4 0 0},clip,width=0.9\columnwidth]{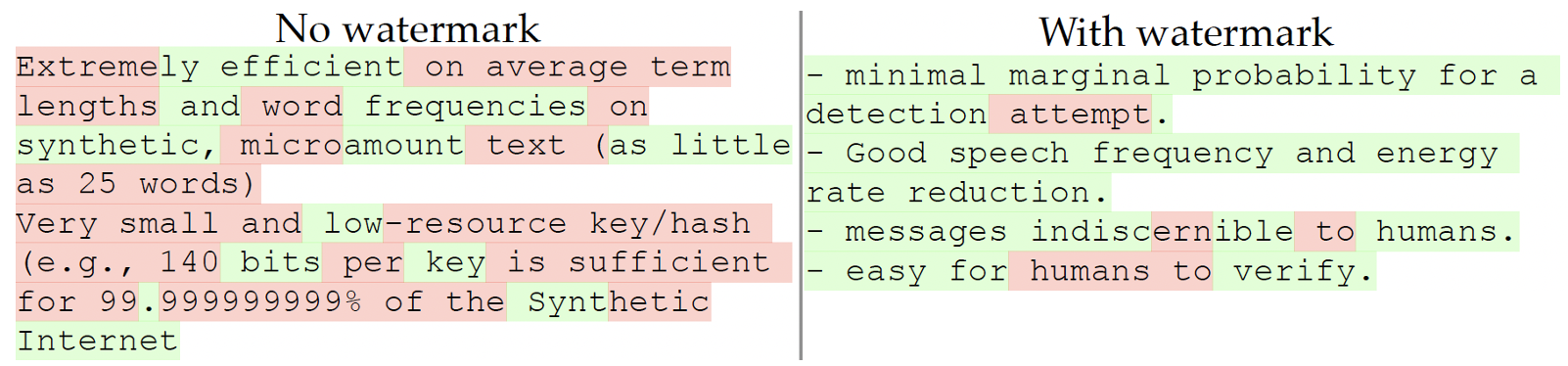}
\caption{Input-encrypted input pair after adding the text templates into the original text. The technique samples the text more from the green tokens list for encryption~\citep{kirchenbauer2023watermark}.}
\label{fig:text_input_enc_input}
\vspace{-3mm}
\end{figure}

\Paragraph{Word Tokens}
Word tokens are commonly used to create templates by replacing words with token words that are close in the embedding space, preserving the original meaning of the sentence.  We show some examples in~\cref{fig:text_input_enc_input}. First comes the usage of the most common form of template: word tokens~\citep{munyer2023deeptextmark, liu2023watermarking}. Creating a list of word tokens involves the generation of candidate words by removing extraneous text elements and converting each word into an embedding vector using a pretrained Word2Vec~\citep{mikolov2013efficient} model. The encryption process is then executed by replacing words with token words that are close in the embedding space~\citep{munyer2023deeptextmark, wang2024learning, wu2022promptchainer, gao2022visual}, preserving the sentence's original meaning as assessed by an encoder. This encoder evaluates the quality of the added template by ensuring the encrypted sentence maintains a high similarity score with the original sentence.~\citet{munyer2023deeptextmark} adopt a logits-based approach. A green token's logit is obtained for each word, then modified logits are passed through a softmax operator to establish a new probability distribution over the vocabulary. This subtly alters the text in a way that embeds a template without significantly changing the text's apparent meaning or readability. Another method by~\citet{he2022cater} inject token words in the conditional word distribution while maintaining the original word distribution. This technique uses linguistic features as a condition for the substitution, which allows the template to be embedded in a way that is sensitive to the text's syntax and semantics.

\begin{figure}[t!]
\centering
\includegraphics[trim={0 -4 0 0},clip,width=\columnwidth]{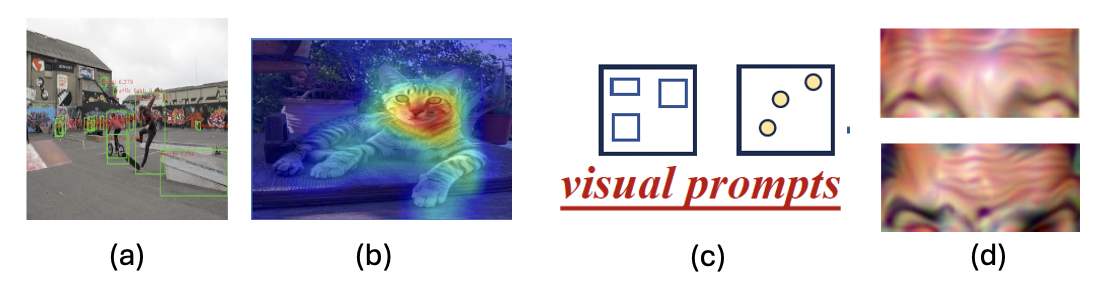}
\caption{Prompts as a type of templates. Techniques include (a) bounding boxes to the images~\citep{girshick2014rich, asnani2024probed}, (b) using attention maps~\citep{li2018tell}, (c) text and visual prompts~\citep{jiang2024t}, and (e) adversarial prompts~\citep{komkov2021advhat}.}
\label{fig:sec_2_prompt_example}
\vspace{-3mm}
\end{figure}

\Paragraph{Character-Level Substitutions}
Character-level substitutions involve replacing or augmenting text with specific characters, punctuation marks, or selected words/sentences to create templates.~\citet{liu2023watermarking} propose a methodology where text is replaced or augmented with triggers, ensuring contextual appropriateness and text integrity. Similarly,~\citep{li2023black, li2020open, monden2000practical} and utilize triggers like black marks or specific patterns appended to images for encryption, aiding in image identification or manipulation detection. These triggers, described as hidden signatures, are added using backdoor attacks to detect unauthorized modifications.~\citep{robey2023smoothllm, dong2023revisit} explore character-level perturbations in prompts for LLMs, where characters are swapped or sampled to integrate perturbations seamlessly with natural text variability. Lastly,~\citet{rizzo2019fine} employ a cryptographic keyed hash function to substitute characters with homoglyphs—visually similar characters with different encodings.

\Paragraph{Text Strings and Masking}
Using text strings and masking techniques involves transforming training data with keys or selectively obscuring information within text data to create templates.~\citet{zhang2018protecting} use text strings, images from other classes, gaussian noise, etc., to use as templates. The original training data is transformed with a key indicating how to label the template. The true label of the data and the predefined label for the template are used to output a protected DNN model and the templates. Masking is also employed for text data to perform encryption. Utilized for masking sentences,~\citep{guo2023authentigpt} employ selective obscuring of sensitive information within textual data. Masks offer a means of controlling the visibility of specific segments of text, providing a mechanism for privacy protection or controlled data disclosure.

In summary, these works present advanced approaches to text templates, utilizing various signal types, such as word tokens and character substitutions, along with complex encryption processes that account for the linguistic and statistical properties of text. These methods aim to embed templates that are challenging to detect and remove while ensuring robustness for reliable verification and attribution.
%Next, we will discuss the works which rely on visual prompts as the type of templates for different applications. 

\begin{table*}[ht]
    {\footnotesize
    \centering
    \begin{tabularx}{\textwidth}{|>{\centering\arraybackslash}m{1.8cm}|>{\centering\arraybackslash}m{4cm}|>{\centering\arraybackslash}m{4.5cm}|>{\tiny\centering\arraybackslash}m{5.54cm}|}
    \hline
    \rowcolor[HTML]{EFEFEF} \textbf{Category} & \textbf{Description} & \textbf{Keywords} & \textbf{References} \\
    \hline
    Visual Prompts & 
    Embedding visual prompts into images or video frames to guide models. & 
    visual prompts, images, videos, pixel level, inference guidance & 
   ~\citep{tsao2023autovp,bahng2022exploring,tsai2024convolutional,wu2022unleashing,zhang2023text,zhu2023visual,sohn2023visual} \\
    \hline

    Advanced Embedding Techniques & 
    Modifying low-frequency components, using siamese architectures, and class-specific prompts. & 
    low-frequency components, siamese networks, class-specific prompts, augmentation & 
   ~\citep{han20232vpt,wang2023fvp,pei2024sa2vp,kunananthaseelan2024lavip,kim2024improving,bar2022visual,chen2023visual,oh2023blackvip,cai2024vip,chen2024vp3d,chen2023understanding,zhang2024visual,kim2024we,wang2024revisiting,yao2024cpt} \\
    \hline

    Multimodal Prompts & 
    Combining text and visual data to enhance model performance. & 
    multimodal prompts, text, visual data, vision-language models & 
   ~\citep{jiang2024t,yang2022prompting,xing2023dual,wen2022visual,long2023fine,lee2023multimodal} \\
    \hline
    \end{tabularx}
    \caption{Summary of works which utilize prompts as the template.}
    \label{tab:summary_prompt_sec2}}
\end{table*}

\begin{figure}[t!]
\centering
\includegraphics[trim={0 -4 0 0},clip,width=\columnwidth]{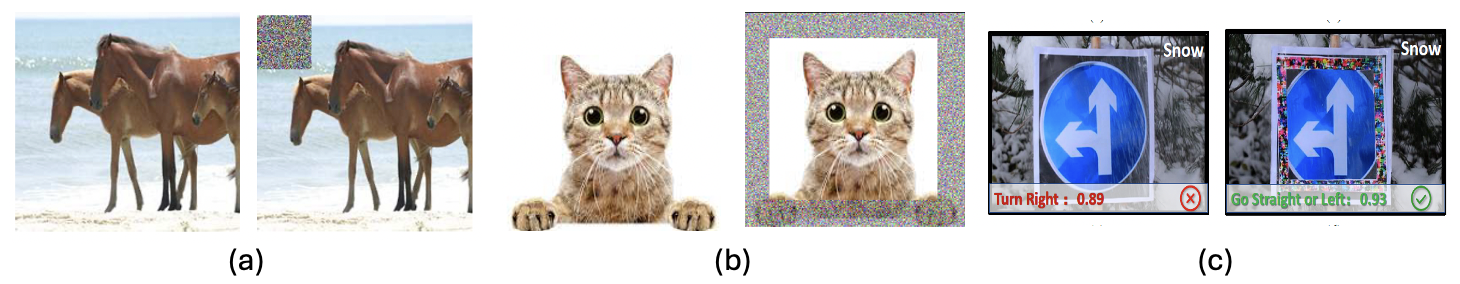}
\caption{Various examples of input-encrypted input pairs after adding prompt templates into the original input images. The templates are added as a patch on a fixed location~\citep{ong2021protecting, li2023visual, komkov2021advhat}, on different locations~\citep{li2023visual}, or on edges for the images~\citep{wang2022defensive, kunananthaseelan2024lavip, li2023visual, yang2023visual}. (a)~\citep{ong2021protecting}, (b)~\citep{yang2023visual}, and (c)~\citep{bahng2022exploring}.}
\label{fig:prompt_input_enc_input}
\vspace{-3mm}
\end{figure}

\subsection{Prompts}
Prompts have been widely used as a proactive technique, appending visual or multi-modal prompts to input images or captions. Visual prompting involves embedding cues directly into image or video data to guide machine learning models, as shown in~\cref{fig:sec_2_prompt_example}. These prompts can be embedded at the pixel level, altering the image data directly. Examples of input-encrypted input pairs are shown in~\cref{fig:prompt_input_enc_input}. Prompts are extensively used in vision and language-based applications. A summary of techniques for embedding prompts is provided in~\cref{tab:summary_prompt_sec2}, with further discussion below.

\Paragraph{Visual Prompts in Images and Videos}
Visual prompts are embedded into input images or video frames to provide additional guidance to models during inference. 
In~\citep{tsao2023autovp, bahng2022exploring, tsai2024convolutional, wu2022unleashing}, visual prompts are embedded into images at the pixel level or as additional channels to guide model inference.~\citet{zhang2023text} apply prompts to video frames, embedding them in the pixel space to alter visual features.~\citet{zhu2023visual} use a visual transformer encoder with 1D tokens of flattened images.~\citet{sohn2023visual} add prompts through prompt tuning, optimizing token generator parameters while keeping transformer models fixed. Various visual prompts, such as strokes, masks, boxes, scribbles, and points, are embedded and extracted using a prompt encoder.

\Paragraph{Advanced Embedding Techniques}
Innovative methods for embedding visual prompts include modifying low-frequency components, using siamese architectures, and augmenting input images with class-specific prompts. Various works adopt innovative ways to embed visual prompts into images. For instance,~\citet{han20232vpt} add visual and key-value prompts during fine-tuning, inserting them into each transformer layer’s input sequence and self-attention module, respectively.~\citet{wang2023fvp} alter the low-frequency components of an image in the frequency domain to add the visual prompt, modifying the amplitude and phase components with learnable parameters.~\citet{pei2024sa2vp} adopt a siamese dual-pathway architecture to embed and extract the prompts using separate pathways, aligning spatially with image tokens to capture detailed information. Other works~\citep{kunananthaseelan2024lavip, kim2024improving} embed visual prompts into the model by augmenting input images with class-specific visual prompts.~\citet{bar2022visual} use grid-like images containing input-output pairs and query images with prompts for inpainting applications. Visual prompts are also referred to as different types of perturbations, images~\citep{chen2023visual, oh2023blackvip, cai2024vip, chen2024vp3d, chen2023understanding} or vectors~\citep{zhang2024visual}, which are added directly to the images and to the activation maps within the model. Works like~\citep{kim2024we, wang2024revisiting} and~\citep{yao2024cpt} treat the prompts as learnable tokens and colored blocks, respectively.

\iffalse
\begin{figure}[t!]
\centering
\includegraphics[trim={0 -4 0 0},clip,width=\columnwidth]{figures/video_method.png}
\caption{Method using templates for video applications (a) adding template to the video using a network~\citep{luo2023dvmark}, (b) adding vidual prompts to all the frames of the video~\citep{zhang2023text}.}
\label{fig:sec_3_video_method}
\vspace{-3mm}
\end{figure}
\fi

\Paragraph{Multimodal Prompts}
Multimodal prompts combine text and visual data to enhance model performance~\citep{jiang2024t, yang2022prompting}. For instance, in~\citep{yang2022prompting}, templates are embedded by applying prompts and perturbations to input videos at frame-level, modifying data distribution for enhanced tracking performance.~\citet{xing2023dual} and~\citet{wen2022visual} propose that text and visual prompts are embedded using pretrained vision-language models, with text prompts processed by the text encoder and visual prompts inserted into the image encoder.

In summary, visual prompts are embedded into image or video data at the pixel level, in frame-aware sequences, or within transformer models via low-frequency modifications or activation map perturbations. Multimodal prompts combine text and images using vision-language models, optimized through prompt engineering and model inversion to enhance model performance while maintaining data integrity.

\subsection{Others}
\begin{figure}[t!]
\centering
\includegraphics[trim={0 -4 0 0},clip,width=\columnwidth]{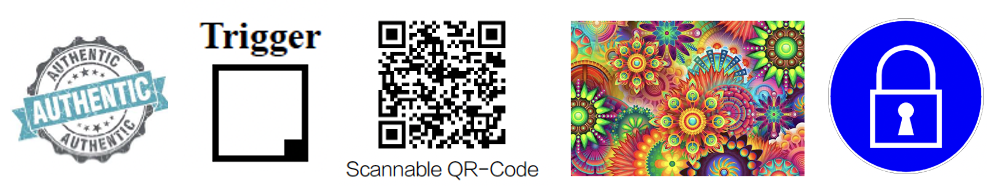}
\caption{Other type of templates. These include multiple tags for different purposes~\citep{meng2022traceable, sun2023faketracer}, authorship rights~\citep{wu2020watermarking}, images~\citep{wu2020watermarking}, (d) triggers, qr codes, predefined templates~\citep{li2023black, zhao2023recipe, kapusta2023protecting,wang2023plug}, and predefined images~\citep{adi2018turning}.}
\label{fig:sec_2_other_example}
\vspace{-3mm}
\end{figure}

\begin{figure}[t!]
\centering
\includegraphics[trim={0 -4 0 0},clip,width=\columnwidth]{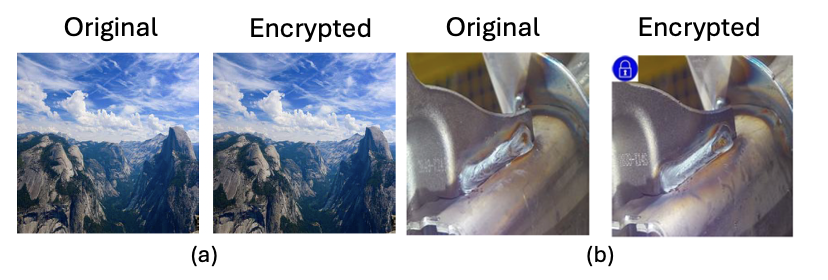}
\caption{Various examples of input-encrypted input pairs after adding tag templates into the original input images. (a)~\citep{wang2023plug} and (b)~\citep{kapusta2023protecting}.}
\label{fig:tag_input_enc_input}
\vspace{-3mm}
\end{figure}

Apart from the main types of templates discussed above, there are several other types of templates that are adopted by various works. These involve tags, qr codes, images, etc. Some of the examples suing these templates and the input-encrypted input are shown in~\cref{fig:sec_2_other_example} and~\cref{fig:tag_input_enc_input}, respectively. We show the summary of these techniques in~\cref{tab:summary_other_sec2}, and discuss these below.

\Paragraph{Differential Excitation and Audio Signals}
Differential excitation and audio signal templates provide robust encryption by embedding templates based on image content or audio alterations.~\citet{laouamer2015robust} propose using the Weber differential excitation descriptor to embed a template into an image by computing the differential excitation for each block, ensuring the template is tied to the image characteristics and resilient to alterations.
Other encryption methods involve adding template signals to audio data.~\citet{xu2021audio} convert audio signals into noise for encryption, altering the original audio content while maintaining its format, making it difficult for unauthorized users to extract meaningful information.

\Paragraph{Random Noise and Predefined Templates}
Random Noise encompasses techniques like direct addition of noise to feature spaces or embedding Gaussian noise vectors into mesh coordinates output by $3$D models~\citep{zhu2021gaussian, nakazawa2010visually, medimegh20183d, cho2006oblivious}. For example, k-Same-Pixel~\citep{newton2005preserving} directly operates on pixel values, while Gaussian noise vectors are added to mesh vertex coordinates to create robust templates. The encryption process for Gaussian noise involves adding random noise to high-frequency coefficients, ensuring the templates survive various attacks and maintain robustness against common mesh operations. Techniques like hiding noise in sensitive samples~\citep{liu2021secure, dwork2006differential} and using Gaussian noise to create trigger patterns for backdoor encryption are also prevalent.

Predefined templates are also explored in many works which involve embedding unique identifiers like tags~\citep{wang2016apriltag, olson2011apriltag, garrido2014automatic}, fiducial markers~\citep{romero2019fractal, fiala2005artag} or predefined triggers~\citep{zhang2023black, lim2022protect} into the images or $3$D models~\citep{zhang2024gs,ohbuchi1998watermarking}. These tags are detected using specialized algorithms that identify and localize these markers within the images. The encryption process includes encoding binary payloads into planar markers, which are used for pose estimation and model verification. Other predefined tags are embedded into the models using neural networks~\citep{krogius2019flexible, cao2023invisible} or added into the mesh~\citep{zhang2024gs,ohbuchi1998watermarking} that ensure robust detection and pose estimation capabilities.

\Paragraph{Sinusoidal Signals and Digital Signatures}
Sinusoidal signals and digital signatures are embedded into data distributions and neural networks to enhance security and enable tamper detection. Sinusoidal template signals are used for encrypting data.~\citet{zhao2023protecting} embed sinusoidal signals into data distributions using hash functions, ensuring concealment and integrity, which enhances security and robustness.
Another technique employs Tardos-like fingerprinting with nearest neighbor decoding.~\citet{laarhoven2019nearest} utilize probability-based vectors for fingerprinting and identification, adding security or traceability to data. This facilitates the identification of unauthorized copies or alterations based on statistical similarities in the data distribution.

~\citet{fan2019rethinking} introduce a passport layer after convolutional layers. This approach incorporates digital signatures into neural networks, providing authentication or tamper detection capabilities to model outputs. By appending digital signatures, it enables verification of model predictions and ensures the integrity of model behavior.~\citet{liu2018local} propose estimating spatial chaotic maps to the prior encryption methods. To improve security, the spatiotemporal chaotic system is widely applied to chaotic cryptography because of its improved chaotic dynamic performance.

\begin{table*}[ht]
{\footnotesize
    \centering
    \begin{tabularx}{\textwidth}{|>{\centering\arraybackslash}m{1.8cm}|>{\centering\arraybackslash}m{3cm}|>{\centering\arraybackslash}m{3.5cm}|>{\tiny\centering\arraybackslash}m{7.54cm}|}
    \hline
    \rowcolor[HTML]{EFEFEF} \textbf{Category} & \textbf{Description} & \textbf{Keywords} & \textbf{References} \\
    \hline
    Differential Excitation and Audio Signals & 
    Embedding templates based on image content or audio alterations. & 
    differential excitation, audio signals, encryption, robust templates & 
   ~\citep{laouamer2015robust,xu2021audio} \\
    \hline

    Random Noise and Predefined Templates & 
    Adding noise to feature spaces or embedding unique identifiers like tags or markers. & 
    random noise, predefined templates, tags, markers, robust templates & 
   ~\citep{zhu2021gaussian,nakazawa2010visually,medimegh20183d,cho2006oblivious,hamidi2017robust,ai2009new,ohbuchi2002frequency,pham20183d,yu2003robust,ohbuchi1998data,bors2006watermarking,alface2007blind,praun1999robust,wang2016apriltag,olson2011apriltag,garrido2014automatic,abbas2019analysis,romero2018speeded,alvarez2012new,wagner2007artoolkitplus,wang2018hierarchical,krogius2019flexible,romero2019fractal,fiala2005artag,zhang2023black,lim2022protect,kapusta2023protecting,ahmadi2020redmark,peng2023intellectual,li2022black,zhang2024gs,ohbuchi1998watermarking,cao2023invisible,ong2021protecting,szyller2021dawn} \\
    \hline

    Sinusoidal Signals and Digital Signatures & 
    Embedding sinusoidal signals and digital signatures to media. & 
    sinusoidal signals, digital signatures, data distribution, neural networks & 
   ~\citep{zhao2023protecting,laarhoven2019nearest,fan2019rethinking,liu2018local} \\
    \hline

    Code Modifications and Transformations & 
    Obscuring the functionality and structure of software programs. & 
    code modifications, transformations, obfuscation, ASCII encoding & 
   ~\citep{monden2000practical,balachandran2014function,li2023functionmarker} \\
    \hline
    \end{tabularx}
    \caption{Summary of works that utilize different forms of templates other than the main ones.}
    \label{tab:summary_other_sec2}}
\end{table*}

\Paragraph{Code Modifications and Transformations}
Code modifications and transformations enhance security by obscuring the functionality and structure of software programs. Techniques such as overwriting numerical operands or replacing opcodes are commonly used in code obfuscation~\citep{monden2000practical}, making it harder for attackers to understand or reverse-engineer the program. Code transformations include function-level obfuscation, where code segments from different functions are shuffled.~\citet{balachandran2014function} use this technique to obscure the logical flow and structure of programs, complicating efforts to analyze or understand their inner workings. These methods enhance security against unauthorized access or tampering.

\citet{li2023functionmarker} encode data using ASCII encoding with a variable length. ASCII encoding is a common technique for converting textual and symbolic data into a standardized format, with the variable length aspect allowing for efficient representation of diverse information. The approach by~\citet{li2023functionmarker} involves injecting a function into a dataset to generate input-output pairs using ASCII encoding, thereby augmenting the available training data for machine learning models. Finally, object detection works~\citep{girshick2014rich, kong2016hypernet, brazil2019m3d, ren2016faster} heavily rely on region proposals in the image representing possible bounding boxes for objects in the image. These region proposals can be considered as a type of template which is added to the image before passing it through the detection framework.

In conclusion, various other templates utilize techniques such as differential excitation, audio signal encryption, random noise addition, predefined tags, sinusoidal signals, digital signatures, and code modifications. These methods are crafted to embed templates that are hard to detect and remove while ensuring robust verification and attribution. Each technique offers distinct advantages in security, robustness, and applicability, making them suitable for a wide range of data types and applications.

\section{Template Learning}
The learning process for embedding templates, such as bit sequences, $2$D templates, text signals, prompts, and others, involves integrating these templates into digital content with minimal visual or functional impact. Different types of templates have different learning paradigm, and various metrics are used to evaluate the learning of template. A summary of all the works is given in~\cref{tab:summary_bit_seq_sec3}. We now will discuss the learning process employed by various types of templates.

\subsection{Bit Sequences}

In proactive learning, bit sequence embedding is crucial for enhancing security and verifying digital content. Various methods integrate bit sequences into data, ensuring integrity and authenticity through structured encoding and decoding processes. The encoder-decoder framework forms the backbone of these techniques, while advanced neural network techniques leverage innovative network architectures. Additionally, advanced techniques in $3$D data ensure secure encryption in $3$D models, collectively enhancing digital content security across domains.

\Paragraph{Encoder-Decoder Framework}
The encoder-decoder models are foundational for embedding and extracting bit sequence templates, ensuring high fidelity and robustness through structured encoding and decoding processes~\citep{zhao2023proactive, sun2023faketracer, yang2021faceguard}, functioning to embed templates by integrating them with the content's identity and subsequently ensuring the fidelity of the template through the decoding process. The efficacy of this method is predominantly evaluated using bit accuracy metrics~\citep{sun2023faketracer, yang2021faceguard, wang2021faketagger}, which ascertain the precision of the added template after potential content manipulation. Some works have further refined this approach by fine-tuning encoder-decoder networks with deepfake models~\citep{sun2023faketracer} or diffusion models~\citep{zhao2023recipe}, thereby enhancing the model's capability to detect and restore content authenticity with greater accuracy, as indicated by lower bit error rates. Some methods which employ encoder-decoder frameworks use statistical analysis like the p-value of the null hypothesis test~\citep{haghighi2018trlh, yu2021artificial}. Multiple encoders are employed by~\citet{wu2020watermarking} to add bit sequence templates into the image while a single decoder is used to extract the added template. Various loss functions are used to guide the training.

\Paragraph{Advanced Neural Network Techniques}
These methods leverage innovative network structures and training techniques to improve bit sequence embedding and extraction, often incorporating adversarial training and statistical analyses. The Inverse Decoupled Invertible Neural Network (DINN)~\citep{meng2022traceable} allows tracing templates back through alterations to restore the original template.~\citet{yang2023towards} adopt a network-based approach with separate modules for embedding and extracting templates, enhancing both bit accuracy and template accuracy.~\citet{zeng2023securing} introduce adversarial training methods where a template injector and classifier embed templates within neural networks. The classifier performs attribution, marking decisions to identify the presence and ownership of templates, useful in intellectual property disputes.

Further,~\citep{uchida2017embedding, fernandez2023stable, nagai2018digital, liu2021watermarking} embed templates by adding a regularization term to the original cost function of the neural network. The decision process involves bit error rate analysis~\citep{uchida2017embedding, nagai2018digital, liu2021watermarking} and statistical hypothesis testing for template verification.~\citet{fernandez2023stable} use public key cryptography for signature extraction and verification, ensuring that only the rightful owner can claim their embedded template. Methods utilizing deep spatial encryption frameworks involve embedding sub-networks to insert templates into target models using additive-based embedding and noise layers~\citep{zhang2021deep}. The extraction relies on trained sub-networks to isolate the template, with metrics like Peak Signal-to-Noise Ratio (PSNR) and Structural Similarity Index (SSIM) used for visual quality evaluation. In federated deep neural networks, feature-based and backdoor-based templates are embedded, with fidelity assessed through classification accuracy and statistical significance~\citep{li2022fedipr}.

\Paragraph{Evaluation Metrics as Objective Functions}
Evaluation metrics such as bit-error ratio, SSIM, PSNR, and LPIPS are critical for assessing the integrity of embedded templates and the overall content quality post-manipulation. Therefore, some techniques utilize these metrics as the objective functions.~\citet{wu2023sepmark} propose an encoder-decoder approach where the evaluation relies on metrics such as bit-error ratio, SSIM, PSNR, and LPIPS. These metrics assess the template's integrity post-manipulation and evaluate the overall content quality, ensuring that encryption does not compromise usability while defending against tampering.~\citet{zhu2018hidden} incorporate image distortion and adversarial loss into the encoder-decoder process to enhance adversarial robustness. This approach is crucial for developing templates that can withstand adversarial attacks.
~\citet{zhao2023proactive} employ cross-correlation between the extracted identity and a predefined template, with auto-correlation assessing the template itself.~\citet{paruchuri2009video} leverage selective embedding into Discrete Cosine Transform (DCT) coefficients, aiming to minimize distortion in video data.

\begin{table*}[ht]
{\footnotesize
    \centering
    \begin{tabularx}{\textwidth}{|>{\centering\arraybackslash}m{2.3cm}|>{\centering\arraybackslash}m{3.5cm}|>{\centering\arraybackslash}m{3.5cm}|>{\tiny\centering\arraybackslash}m{6.54cm}|}
    \hline
    \rowcolor[HTML]{EFEFEF} \textbf{Category} & \textbf{Description} & \textbf{Metrics} & \textbf{References} \\
    \hline
    Encoder-Decoder Framework & 
    CNN models for embedding and extracting bit sequence template & 
    Bit accuracy, bit error rates, p-value, SSIM, PSNR, LPIPS & 
   ~\citep{zhao2023proactive,sun2023faketracer,yang2021faceguard,yu2021artificial,wang2021faketagger,wu2023sepmark,cui2023diffusionshield,neekhara2022facesigns,zhang2023editguard,haghighi2018trlh,wu2020watermarking,asnani2024promark} \\
    \hline

    Advanced Neural Network Techniques & 
    Leveraging neural networks and training techniques. & 
    Bit error rate, statistical hypothesis testing, PSNR, SSIM, accuracy & 
   ~\citep{meng2022traceable,yang2023towards,zeng2023securing,uchida2017embedding,fernandez2023stable,nagai2018digital,liu2021watermarking,zhang2021deep,li2022fedipr} \\
    \hline

    Evaluation Metrics as Objective Functions & 
    Using metrics as objective functions. & 
    Bit-error ratio, SSIM, PSNR, LPIPS, image distortion& 
   ~\citep{wu2023sepmark,zhu2018hidden,zhao2023proactive,paruchuri2009video} \\
    \hline

    $3$D Domain & 
    Vertex modifications in point clouds and SDFs. & 
    Bit accuracy, MRMS, HD, RMSC, correlation coefficient, SNR & 
   ~\citep{chen2024nerf,jang2024waterf,luo2023copyrnerf,zhu2023towards,mun2015robust,kanai1998digital,uccheddu2004wavelet,wang2008hierarchical,kim2005watermarking,liu2019novel,al2019graph,ohbuchi2001watermarking,peng2022semi,zhu2024rethinking} \\
    \hline

    Miscellaneous Techniques & 
    Techniques for IP protection, embedding digital signatures. & 
    Bit accuracy, classification accuracy, semantic resemblance & 
   ~\citep{chen2019deepmarks,darvish2019deepsigns,yang2022tracing} \\
    \hline
    \end{tabularx}
    \caption{Summary of encryption techniques used for adding bit sequence as template.}
    \label{tab:summary_bit_seq_sec3}}
\end{table*}

\Paragraph{3D Domain}
Techniques for embedding templates in $3$D data include methods for modifying vertex distributions and embedding binary messages in point clouds and Signed Distance Fields (SDFs). templates are integrated into the rendering process or embedded using Implicit Neural Representation (INR) and specific keys~\citep{chen2024nerf, jang2024waterf, luo2023copyrnerf}. For SDFs~\citep{zhu2023towards, mun2015robust}, binary template messages are embedded through local deformations within spherical partitions. The extraction from templateed SDFs is evaluated based on bit accuracy. Further, wavelet analysis\citep{kanai1998digital, uccheddu2004wavelet, wang2008hierarchical, kim2005watermarking} is used to obtain approximation meshes and wavelet coefficients, with salient points extracted based on mesh saliency. These methods are evaluated using metrics like Mean Root Mean Square (MRMS) and Hausdorff Distance (HD).

Techniques involving point clouds calculate the Root Mean Square Curvature (RMSC) values of vertices and establish synchronization relations to embed template information~\citep{liu2019novel}. The extraction process calculates the correlation coefficient between the extracted and original templates.~\citet{al2019graph} use the Graph Fourier Transform (GFT) to embed templates in sorted GFT coefficients, with extraction relying on these coefficients and specific selection conditions. Spectral domain methods modify mesh spectral coefficients to embed templates resistant to transformations and noise~\citep{ohbuchi2001watermarking}, evaluating metrics like perceptibility, robustness, and resistance to disturbances. Variable Direction Double Modulation (VDDM)~\citep{peng2022semi} transforms $3$D models into spherical coordinates, embedding templates based on vertex positions in the one-ring neighborhood. Extraction involves recovering templates from encrypted and plaintext domains, assessed using imperceptibility and bit error rate (BER). In attention-based method by~\citet{zhu2024rethinking}, vertex distributions are modified based on binary messages. The extraction involves decoding binary messages from templateed vertices, with metrics like Hausdorff distance and signal-to-noise ratio (SNR) used to measure geometric distortion. Advanced methods embed templates by perturbing $3$D point coordinates or modifying vertex norm histograms~\citep{mun2015robust}.

\begin{table*}[ht]
{\footnotesize
    \centering
    \begin{tabularx}{\textwidth}{|>{\centering\arraybackslash}m{1.8cm}|>{\centering\arraybackslash}m{3cm}|>{\centering\arraybackslash}m{3.5cm}|>{\tiny\centering\arraybackslash}m{7.54cm}|}
    \hline
    \rowcolor[HTML]{EFEFEF} \textbf{Category} & \textbf{Description} & \textbf{Metrics} & \textbf{References} \\
    \hline
    Adversarial Perturbations & 
    Optimizing small perturbations to input data. & 
    Accuracy, MSE, SSIM, FDFR, ISM, SER-FQA, BRISQUE, FID, L2 error & 
   ~\citep{ducoffe2018adversarial,xu2023adversarial,xue2022advparams,wu2024watermarks,tang2024once,zhang2021proactive,zhang2022adversarial,shi2017evasion,liu2020adversarial,kitada2021attention,dong2019efficient,van2023anti,huang2022cmua,segalis2020ogan,ruiz2020disrupting,yeh2020disrupting,zhong2020towards,yu2019attributing,li2019anonymousnet,agrawal2000privacy,meng2017magnet,ye2023duaw,he2023perturbation} \\
    \hline

    Learnable Perturbations & 
    Estimating perturbations using task-specific loss functions. & 
    PSNR, SSIM, LPIPS, template Detection Rate, False Positive Rate & 
   ~\citep{asnani2024probed,Asnani_2022_CVPR,Asnani_2023_CVPR,wong2020learning,wang2023plug,tekgul2021waffle,wang2022defensive,cui2023ft,zhao2024can,xiao2021improving,peng2022fingerprinting,cozzolino2019noiseprint} \\
    \hline

    GANs & 
    GANs estimate templates. & 
    ASR, PSNR, SSIM, KL divergence & 
   ~\citep{hu2022protecting,xiong2020adgan,wu2019privacy} \\
    \hline

    Privacy Preservation & 
    Minimizing classifier accuracy to protect identities. & 
    Genuine/imposter match scores, perceptual similarity & 
   ~\citep{mirjalili2018semi,shan2020fawkes,othman2015privacy,cherepanova2021lowkey} \\
    \hline

    Geometric Perturbations & 
    Displacing triangle medians in $3$D models. & 
    Imperceptibility, capacity, mean distortion, L0, L1, L2 norms & 
   ~\citep{molaei2013blind,dhillon2018stochastic,li2024deep} \\
    \hline
    \end{tabularx}
    \caption{Summary of encryption techniques used for adding $2$D noises as template. }
    \label{tab:summary_2d_sec3}}
\end{table*}

\Paragraph{Miscellenaous Techniques}
Some techniques for IP protection include embedding digital signatures in neural networks and fine-tuning weights to trigger specific templates.~\citet{chen2019deepmarks} proposes to acquire the weights in the marked layers to reconstruct the class-specific FP vector which is then correlated by the predicted score vector for IP protection.~\citet{darvish2019deepsigns} also fine-tune the weights of the neural network by creating specific input keys to later trigger the corresponding template and use the recovered template for the task of IP protection.

Text-based methods use semantic models to encrypt content, ensuring the original message's meaning is preserved. Some works~\citep{yang2022tracing} employ semantic models like BERT~\citep{devlin2018bert} for encryption, maintaining the message's meaning. The learning objective involves semantic analysis, ensuring the encrypted content resembles the original sentence in meaning, beyond just statistical measures.

In summary, bit sequence embedding techniques leverage the encoder-decoder framework and advanced network architectures ensure robust and accurate template embedding. Comprehensive evaluation metrics and neural network-based methods further strengthen this learning process of the bit-sequence templates. Collectively, these approaches provide a robust framework for maintaining the integrity and reliability of digital content across various applications. Next, we list out works involving learning process for $2$ templates.

\subsection{$2$D templates}
Learning the class of $2$D templates has progressed significantly, incorporating various perturbation techniques to ensure robust and secure encryption and improvement in various applications. These methods use adversarial perturbations, learnable perturbations, and geometric transformations to embed template signals effectively. A summary of all the works is given in~\cref{tab:summary_2d_sec3}. We will now discuss these methods in details. 

\iffalse
\begin{figure*}[t!]
\centering
\includegraphics[trim={0 -4 0 0},clip,width=\textwidth]{figures/bit_seq_method_2.png}
\caption{Bits sequence learning. Some of the techniques include (a) adding bit to text inputs via hash~\citep{rizzo2019fine}, (b) adding to various parts of images~\citep{kiatpapan2015image},     (c) wavelet transform~\citep{lu2001multipurpose}, and (d) using a network~\citep{zhang2023editguard}.}
\label{fig:sec_3_bit_seq_method}
\vspace{-1mm}
\end{figure*}
\fi

\Paragraph{Adversarial Perturbations}
Adversarial perturbations involve optimizing small but intentional perturbations to input data, misleading models while maintaining imperceptibility to human observers. 
Adversarial perturbations, optimized using techniques like Fast Gradient Sign Method (FGSM), Basic Iterative Method (BIM), Carlini and Wagner (C\&W), Iterative Fast Gradient Sign Method (I-FGSM), and Projected Gradient Descent (PGD), are integrated into neural network feature spaces. The perturbations are estimated through classifiers, aiming to mislead models while maintaining imperceptibility~\citep{ducoffe2018adversarial, xu2023adversarial, xue2022advparams, wu2024watermarks, tang2024once}. The evaluation metrics for these methods often include classification accuracy and robustness measures such as mean squared error (MSE) and structural similarity indices. 

Adversarial attack methods create benign adversarial examples to either distort a model's output~\citep{van2023anti, huang2022cmua} or benefit a victim model~\citep{yang2021towards, zhong2020towards}. These methods introduce small perturbations that are imperceptible to humans but cause machine learning models to make mistakes, often for social good. Different optimization processes, such as various attacks and loss functions, estimate learnable perturbations. The method uses metrics like Face Detection Failure Rate (FDFR), Identity Score Matching (ISM), SER-FQA, BRISQUE, Frechet Inception Distance (FID), and L$2$ distance to assess the effectiveness of perturbations.~\citet{yu2019attributing} propose estimating model fingerprints for attributing the source model of an image using neural networks as fingerprints.
Classifiers and encoders are integral in decision-making when assessing perturbations' impact~\citep{li2019anonymousnet, agrawal2000privacy}, taking adversarial perturbations as inputs for applications such as model attribution or computing similarity scores between image fingerprints.

\Paragraph{Learnable Perturbations}
Learnable templates involve estimating a perturbation using task-specific loss functions. The gradients of the methods are backpropagated to update the parameters of these learnable templates to find minimal perturbations improving the respective task~\citep{asnani2024probed, Asnani_2022_CVPR, Asnani_2023_CVPR, jiang2023evading}. These methods are evaluated using PSNR (Peak Signal-to-Noise Ratio), SSIM (Structural Similarity Index), and LPIPS (Learned Perceptual Image Patch Similarity). In the fine-tuning process, methods~\citep{asnani2024probed, Asnani_2022_CVPR, Asnani_2023_CVPR, wong2020learning, huang2024noise, cui2023ft} embed templates by optimizing perturbations to minimize loss while maintaining image quality. Metrics such as the template Detection Rate (TPR) and False Positive Rate (FPR) evaluate the effectiveness of the template embedding.

In perturbation optimization, some approaches refine adversarial perturbations using manifold optimization techniques to create patches that mimic human facial features, enhancing transferability across different models~\citep{xiao2021improving}.~\citet{rajabi2021practicality} and~\citet{peng2022fingerprinting} estimate universal perturbations using encoder networks and ensembles of small CNNs, maintaining effectiveness across images of varying resolutions. Siamese networks are also utilized for template estimation, leveraging dual networks to extract and learn discriminative features from image noiseprints~\citep{cozzolino2019noiseprint}. These networks are particularly useful in forensic applications, where determining the source camera model of an image is necessary. They are trained to recognize subtle noise patterns unique to specific camera models, enabling accurate attribution of images to their origin devices.

\Paragraph{Generative Adversarial Networks (GANs)}
Generative Adversarial Networks (GANs) play a central role in several applications, including estimating learnable templates. For instance,~\citet{hu2022protecting} propose ATM-GAN for transferring makeup styles between images to estimate templates and add those templates to images processed by PP-GAN for de-identifying faces to protect privacy. These GAN architectures are trained with adversarial examples to withstand attacks and fulfill specific image manipulation detection applications. Another example is~\citet{xiong2020adgan}, who use a GAN-based architecture to estimate adversarial examples and disrupt the target model.~\citet{wu2019privacy} propose leveraging GAN-based architecture and contrastive loss to deidentify the facial identity of the images in the dataset. These methods utilize the quality of an image and metrics like Attack Success Rate (ASR), Peak Signal-to-Noise Ratio (PSNR), and Structural Similarity Index (SSIM) to verify the attack on the victim model. These measures are vital for determining the degree to which an image has been compromised by an attack and the perceptual quality of the image compared to its original state. 

\begin{table*}[ht]
{\footnotesize
    \centering
    \begin{tabularx}{\textwidth}{|>{\centering\arraybackslash}m{1.8cm}|>{\centering\arraybackslash}m{6.2cm}|>{\centering\arraybackslash}m{3.5cm}|>{\tiny\centering\arraybackslash}m{4.34cm}|}
    \hline
    \rowcolor[HTML]{EFEFEF} \textbf{Category} & \textbf{Description} & \textbf{Metrics} & \textbf{References} \\
    \hline
    Network-based Methods & 
    Using neural networks in embedding text perturbations for encryption. & 
    Binary detection accuracy, z-statistic, language-level metrics & 
   ~\citep{munyer2023deeptextmark,kirchenbauer2023watermark,dong2023revisit} \\
    \hline

    Perturbation Techniques & 
    Modifying text prompts to enhance model robustness and detect adversarial manipulations. & 
    Similarity metrics, attack resistance, detection accuracy & 
   ~\citep{robey2023smoothllm,li2023functionmarker} \\
    \hline

    Pattern Alterations & 
    Modifying original text using specific patterns to embed templates. & 
    Hash function accuracy, trigger effectiveness, similarity scores & 
   ~\citep{rizzo2019fine,he2022protecting,guo2023authentigpt,sadasivan2023can,krishna2024paraphrasing} \\
    \hline

    Linguistic Features & 
    Creating and verifying templates using linguistic features and statistical testing. & 
    Statistical testing, language metrics, attack success rates & 
   ~\citep{he2022cater,zhang2018protecting,liu2023watermarking} \\
    \hline
    \end{tabularx}
    \caption{Summary of encryption techniques used for adding texts as template.}
    \label{tab:summary_text_sec3}}
\end{table*}

\Paragraph{Privacy Preservation}
Privacy preservation methods aim to reduce the accuracy of specific classifiers, such as gender classifiers, while maintaining high biometric matching accuracy to protect individual identities. For example,~\citet{mirjalili2018semi} reduce gender classifier accuracy, confusing algorithms without affecting biometric matching. This approach in adversarial machine learning ensures privacy by misleading classifiers while preserving identity recognition. Similarly,~\citet{shan2020fawkes} manipulate the feature extraction process in image recognition, using adversarial techniques to alter images and cause models to mislabel them, enhancing privacy protection.
In face morphing applications, templates are estimated using automated gender classifiers, which assign labels and confidence values to quantify gender suppression in images~\citep{othman2015privacy}. The evaluation metrics for these techniques include genuine and imposter match scores, which assess the effectiveness of the perturbations. The LowKey method~\citep{cherepanova2021lowkey} manipulates potential gallery images so that they do not match probe images of the same person. It achieves this by creating a perturbed image whose feature vector is significantly different from that of the original image. 

\Paragraph{Geometric Perturbations}
Geometric perturbations are used as templates by displacing triangle medians in $3$D models~\citep{molaei2013blind}. The extraction process involves analyzing the displacement vectors, with metrics such as robustness, imperceptibility, and capacity. Mean distortion measures the average difference between the original and templateed models. $2$D templates are added to pretrained models using random noise or weight pruning techniques~\citep{dhillon2018stochastic, li2024deep}, creating adversarial examples to test model robustness. These methods are evaluated using norms like L0, L1, and L2 to measure reconstruction error and probability divergence.

Overall, these techniques utilize a combination of perturbation methods and advanced learning models to ensure secure and robust proactive learning. The ongoing advancements promise further improvements in the protection and verification of digital ownership in data, $2$D models, $3$D models, etc., with comprehensive evaluation metrics ensuring the effectiveness of these methods. %We will next discuss learning techniques for texts templates. 

\subsection{Text Templates}
Using text templates is a prominent method in proactive learning, focusing on network based learning, perturbation techniques, and pattern alterations to embed and verify templates in textual data. A summary of all the works is given in~\cref{tab:summary_text_sec3}. We will now discuss these methods in details. 

\Paragraph{Network-based Methods}
Network-based methods involve using neural networks in embedding text perturbations for encryption.~\citet{munyer2023deeptextmark} propose to train a classifier with both encrypted and non-encrypted data. A transformer classifier is used to perform the binary detection. In~\citep{kirchenbauer2023watermark}, the model distinguishes between `green' and `red' tokens, prioritizing the use of `green' tokens especially when the word's entropy is high. This approach is complemented by denoising techniques and statistical transformations to ensure similarity between original and processed sentences. On the decision side,\citep{munyer2023deeptextmark} employs a transformer classifier for binary detection, and\citep{kirchenbauer2023watermark} uses a null hypothesis to determine if a text sequence was generated without knowledge of certain rules. A significant z-statistic leads to the rejection of the null hypothesis, indicating potential machine generation. Sometimes, methods also employ language-level metrics like typos, verbosity, speech, simplification, etc., to evaluate their approach~\citep{dong2023revisit}.

\Paragraph{Perturbation Techniques}
Perturbation techniques involve modifying text prompts to enhance model robustness and detect adversarial manipulations.
Perturbation techniques are applied repeatedly to prompts in the learning process by~\citet{robey2023smoothllm}, with outputs aggregated to resist attacks.~\citet{li2023functionmarker} design multiple template functions with various coefficients, creating input-output pairs and fine-tuning the language model. This robust embedding method is assessed using similarity metrics between original and denoised outputs, helping to determine if the text is machine-generated and evaluating the model's resistance to adversarial manipulation.

\Paragraph{Pattern Alterations}
Pattern alteration methods modify original text using specific patterns to embed templates while preserving semantic integrity.
Some methods alter original text using patterns.~\citet{rizzo2019fine} scan text for confusable symbols and replace them with homoglyphs based on template bits, invisible to readers but detectable in technical analysis.~\citet{he2022protecting} use lexical knowledge to embed semantics-preserving templates, ensuring encryption without compromising text meaning. Trigger functions detect templates for ownership verification.~\citet{guo2023authentigpt} mask and denoise text sentences with a diffusion model, assessing similarity to determine if text is human or machine-generated.~\citet{sadasivan2023can} use recursive paraphrasing to train data security methods with paraphrased text pairs.~\citet{krishna2024paraphrasing} fine-tune LLMs with paraphrased text data to produce text with a similar paraphrasing style.

\Paragraph{Linguistic Features}
Using linguistic features and statistical testing, these methods create and verify robust, semantics-preserving templates in text.~\citet{he2022cater} employ part-of-speech tags and dependency trees to generate templates resilient to text generation API manipulations, evaluating their accuracy through statistical testing and p-values.~\citet{zhang2018protecting} design prompts for verifying AI service ownership, while~\citet{liu2023watermarking} fine-tune LLMs with encrypted data for IP protection, using language-level metrics and hypothesis testing to detect and extract templates with high robustness.

In summary, proactive learning uses techniques like network detection, perturbation, pattern alterations, and linguistic encryption to secure text data, ensuring integrity and protection against adversarial attacks and unauthorized manipulation.%Next, we provide a summary for methods using prompts as the templates. 

\subsection{Prompts}
The extraction of visual and text prompts from embedded signals in machine learning models involves a variety of sophisticated processes and metrics to ensure accurate and effective signal utilization. These prompts can be visual, text-based, or a combination of both, and they are embedded and extracted using various techniques tailored to specific applications and models. A summary of all the works is given in~\cref{tab:summary_prompt_sec3}. We will now discuss these works in details. 

\begin{table*}[ht]
{\footnotesize
    \centering
    \begin{tabularx}{\textwidth}{|>{\centering\arraybackslash}m{1.8cm}|>{\centering\arraybackslash}m{6cm}|>{\centering\arraybackslash}m{4.5cm}|>{\tiny\centering\arraybackslash}m{3.54cm}|}
    \hline
    \rowcolor[HTML]{EFEFEF} \textbf{Category} & \textbf{Description} & \textbf{Metrics} & \textbf{References} \\
    \hline
    Embedded Visual Prompts & 
    Embedding visual prompts into the input space of models for applications like segmentation and classification. & 
    Accuracy, precision, recall, F1 score, Jaccard F-measure (JF), average precision & 
   ~\citep{park2024fair,li2024visual,wu2022promptchainer} \\
    \hline

    Text/Visual Features & 
    Extracting both text and visual features using specific encoders to enhance multi-modal applications. & 
    Accuracy gain, IoU, pixel accuracy & 
   ~\citep{zhu2023visual,wu2022unleashing,tsao2023autovp,kunananthaseelan2024lavip} \\
    \hline

    Generative Vision Transformers & 
    Using prompt tuning in generative vision transformers to optimize and learn prompts. & 
    Accuracy, parameter efficiency, FID & 
   ~\citep{sohn2023visual,han20232vpt} \\
    \hline

    Frequency Domain & 
    Embedding visual prompts in the frequency domain to enhance model robustness and accuracy. & 
    Dice coefficient, Average Surface Distance (ASD), standard accuracy, robust accuracy & 
   ~\citep{wang2023fvp,chen2023visual,bahng2022exploring} \\
    \hline

    Video and Tracking Prompts & 
    Embedding prompts and perturbations in video frames to improve tracking models. & 
    Precision, recall, F-score, Cross-Entropy loss & 
   ~\citep{yang2022prompting,pei2024sa2vp} \\
    \hline

    Advanced Prompt Techniques & 
    Using sophisticated techniques like Colorful Prompt Tuning and Vision-Language Pre-Training Models. & 
    Grounding accuracy, recall@N, mean recall@N, APCER, BPCER, ACER, HTER & 
   ~\citep{yao2024cpt,wen2022visual,yu2023visual} \\
    \hline
    \end{tabularx}
    \caption{Summary of encryption techniques used for adding prompts as template.}
    \label{tab:summary_prompt_sec3}}
\end{table*}

\Paragraph{Embedded Visual Prompts}
Embedding visual prompts into the input space of models helps in fine-tuning and improving applications such as segmentation and classification.
Visual prompts are commonly embedded into the input space of models. For example,\citep{park2024fair} propose that in the Vision Transformer (ViT)~\citep{vaswani2017attention} model, prompts are added to the input sequence during fine-tuning, with mechanisms like Multi-head Self-Attention (MSA) and Masked Multi-head Self-Attention (MSA*) encoding the prompts. In segmentation applications, visual prompts are processed through a prompt encoder to extract meaningful features. These features guide the segmentation process by being combined with input image features and passed through a decoder to generate segmentation masks\citep{li2024visual}. Metrics such as accuracy, precision, recall, and F1 score measure the models' ability to correctly classify or predict target labels~\citep{park2024fair, wu2022promptchainer}. In segmentation applications, metrics like Jaccard and F-measure (JF) and global average precision are employed to compare predicted masks with ground truth masks~\citep{li2024visual}.

\Paragraph{Combined Text and Visual Features}
Using specific encoders to extract both text and visual features improves applications like tracking and classification in multi-modal scenarios.
In scenarios involving both text and visual features, specific encoders extract these features: a trainable word embedding layer for text and a vision encoder like ResNet-50 for video frames. Frame-aware visual prompts are embedded into video frames to enhance applications like tracking and classification~\citep{zhang2023text}. Transformer-based models~\citep{zhu2023visual, wu2022unleashing} use transformer encoder layers for feature extraction and interaction. Auto Visual Prompting~\citep{tsao2023autovp} adds prompts as additional channels using binary masks, extracted with binary classification loss, and evaluated with metrics such as accuracy gain, IoU, and pixel accuracy.

\begin{figure}[t!]
\centering
\includegraphics[trim={0 -4 0 0},clip,width=\columnwidth]{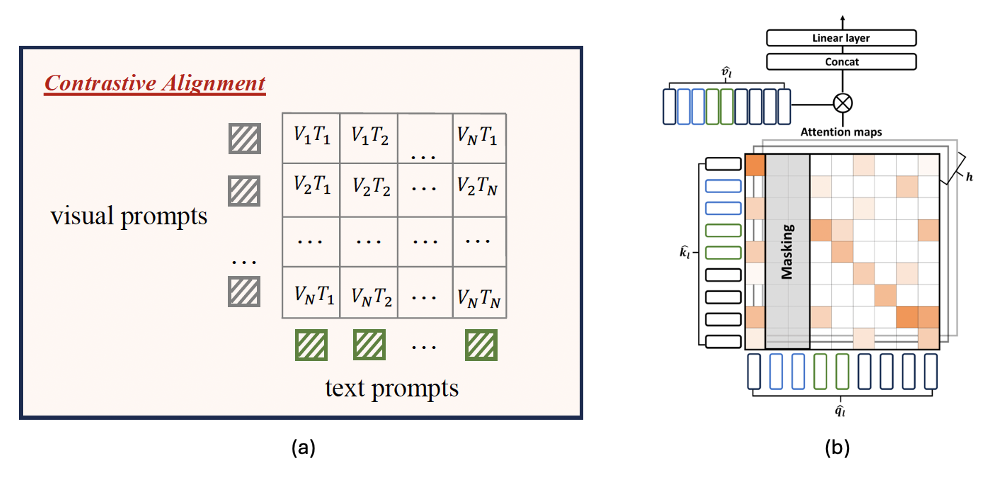}
\caption{Attention-based learning. Attention maps are utilized by either employing (a) contrastive alignment between text and visual prompts~\citep{jiang2024t}, (b) applying masking to attention maps~\citep{park2024fair}.}
\label{fig:sec_3_attention_method}
\vspace{-3mm}
\end{figure}

Further, visual prompts are also extracted by combining language and image-dependent encodings. For instance, in~\citep{kunananthaseelan2024lavip}, the language encoding matrix from a pretrained model is projected and modulated with image-dependent encodings, which are then combined to form visual prompts. This method is integral to models that rely on textual descriptions for class representations, which are crafted and encoded to provide guidance during inference.

\Paragraph{Generative Vision Transformers}
Generative vision transformers use prompt tuning to optimize and learn prompts, enhancing model performance through refined feature extraction.
In generative vision transformers, prompts are learned and added through prompt tuning, with token generator parameters optimized via gradient descent~\citep{sohn2023visual}. The extraction process uses Multi-Layer Perceptron Classifiers (MLPC) and Predictors (MLPP) to encode class and sequence position indices, and performance is evaluated using metrics like Fréchet Inception Distance (FID).
End-to-End Visual Prompt Tuning (E2VPT)~\citep{han20232vpt} involves learning prompts during fine-tuning, inserting predefined text tokens into transformer layers. These prompts are extracted using a pruning strategy to remove the least important ones, and performance is measured using accuracy and parameter efficiency, quantifying the number of tunable parameters and prediction correctness.

\Paragraph{Frequency Domain Visual Prompts}
%Embedding visual prompts in the frequency domain and optimizing them enhances the robustness and accuracy of models.
Visual prompts embedded in the frequency domain are extracted using pre-trained models that generate segmentation predictions, with performance evaluated using metrics like the Dice coefficient and Average Surface Distance (ASD)~\citep{wang2023fvp}. For visual prompts added as perturbations to test-time examples, metrics such as standard accuracy and robust accuracy measure the model's predictions on both normal and adversarial examples~\citep{chen2023visual}. The visual prompt signal is embedded into input images by attaching it to each pixel, influencing the semantics of the image during adaptation. In scenarios where visual prompts are directly optimized via backpropagation~\citep{bahng2022exploring}, they are added to input images, and during training, the model maximizes the likelihood of the correct label given the prompted image. The performance of visual prompting is assessed using metrics such as average accuracy across multiple datasets.

\Paragraph{Video and Tracking Prompts}
Embedding prompts and perturbations in video frames enhances the accuracy and effectiveness of tracking models.
In ProTrack~\citep{yang2022prompting}, the authors propose to embed signals by applying prompts and perturbations to input videos, enhancing the discriminative ability of RGB trackers. The performance of these techniques is evaluated using precision, recall, and F-score, which measure the accuracy and effectiveness of tracking results. In contrast, techniques like SA2VP~\citep{pei2024sa2vp} utilize cross-entropy loss to optimize model performance during training, with metrics like accuracy, precision, recall, and F1 score evaluating performance on benchmarks like VTAB-1k.

\begin{figure}[t!]
\centering
\includegraphics[trim={0 -4 0 0},clip,width=\columnwidth]{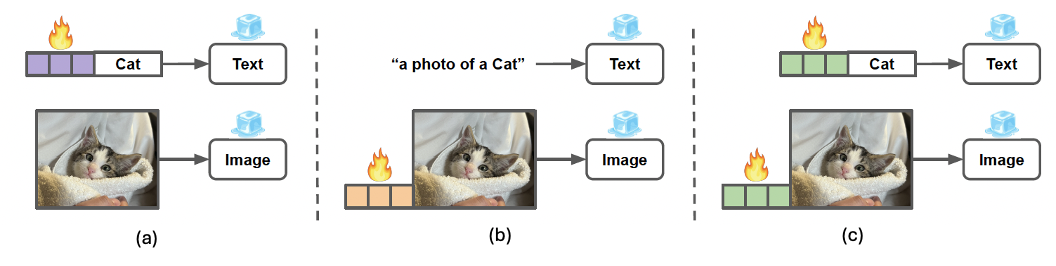}
\caption{Multi-modal prompt learning. Different prompt learning mechanisms shown by~\citep{shen2024multitask, li2023visual, jia2022visual} (a) text prompt trainable, (b) visual prompt trainable, and (c) both text and visual prompt trainable. Other methods include backbone trainable with frozen prompt generator, prompt generator and head is trained with backbone as forzen, \textit{etc.}}
\label{fig:sec_3_prompt_method}
\vspace{-3mm}
\end{figure}

\Paragraph{Advanced Prompt Techniques}
Advanced methods use sophisticated techniques to enhance model performance across various applications.
Advanced methods like Colorful Prompt Tuning (CPT)\citep{yao2024cpt} embed visual sub-prompts by marking image regions with distinct colors or segmentation masks, and textual sub-prompts using color-based query templates. The performance of CPT is evaluated with grounding accuracy, recall@N, and mean recall@N for visual relation detection, as well as accuracy in visual commonsense reasoning and question answering applications. In Vision-Language Pre-Training Models (VL-PTMs)\citep{wen2022visual, yu2023visual}, visual prompts are embedded by optimizing input images through model inversion and the extracted features are evaluated using metrics like Attack Presentation Classification Error Rate (APCER), Bonafide Presentation Classification Error Rate (BPCER), Average Classification Error Rate (ACER), and Half Total Error Rate (HTER).

In conclusion, the use of visual and text prompts in machine learning models enhances performance across various applications. Techniques like combining language and image encodings or embedding prompts in different domains are evaluated using metrics such as accuracy, IoU, FID, and recall. These advancements continue to improve model robustness, accuracy, and functionality.

\subsection{Others}
As discussed earlier, there are some approaches which utilize types of templates different from the above template categories. A summary of all the works is given in~\cref{tab:summary_other_sec3}. We now discuss the learning processes adopted by these works.

\begin{table*}[ht]
{\footnotesize
    \centering
    \begin{tabularx}{\textwidth}{|>{\centering\arraybackslash}m{1.8cm}|>{\centering\arraybackslash}m{5cm}|>{\centering\arraybackslash}m{3.5cm}|>{\tiny\centering\arraybackslash}m{5.54cm}|}
    \hline
    \rowcolor[HTML]{EFEFEF} \textbf{Category} & \textbf{Description} & \textbf{Metrics} & \textbf{References} \\
    \hline
    Encryption in Language Models & 
    Perturbing probability vectors and using sinusoidal signals to embed templates in language models. & 
    PSNR, hypothesis testing, spatial temporal score & 
   ~\citep{zhao2023protecting,li2023black,li2020open,adi2018turning,sablayrolles2020radioactive} \\
    \hline

    Gaussian Noise & 
    Adding Gaussian noise to feature points or vertex coordinates to embed templates in $3$D models. & 
    Detection error, correlation between coefficients, PSNR, SSIM, LPIPS, $\epsilon$ & 
   ~\citep{praun1999robust,zhu2021gaussian,nakazawa2010visually,medimegh20183d,cho2006oblivious,ohbuchi2002frequency,pham20183d,ai2009new,dwork2006differential,zhang2018privacy,cohen2019certified} \\
    \hline

    Trigger and Predefined Tags & 
    Using trigger samples and predefined tags to enhance detection capabilities. & 
    Accuracy, precision, recall, F1 score, trigger recognition rates & 
   ~\citep{zhang2023black,lim2022protect,kapusta2023protecting,ahmadi2020redmark,peng2023intellectual,li2022black,wang2016apriltag,olson2011apriltag} \\
    \hline

    Code Protection and Obfuscation & 
    Ensuring code integrity and protection against unauthorized access. & 
    Robustness, control flow errors, detection range & 
   ~\citep{monden2000practical,balachandran2014function} \\
    \hline

    Fourier Space Encryption & 
    Embedding templates in the Fourier space to provide resilience to various image transformations. & 
    Interpretable P-value, detection threshold ($\alpha$) & 
   ~\citep{wen2023tree} \\
    \hline

    Passport Layers & 
    Adjusting scale factors and using adversarial techniques to enhance model security. & 
    Non-invertibility, detection accuracy, robustness & 
   ~\citep{fan2019rethinking} \\
    \hline

    Frontier Stitching & 
    Clamp the decision frontier. & 
    Statistical framework, hypothesis testing & 
   ~\citep{le2020adversarial,laarhoven2019nearest} \\
    \hline

    Privacy Preservation & 
    Incorporating audio as noise and using chaotic systems to ensure privacy in multimedia content. & 
    Privacy protection level, sensitivity analysis & 
   ~\citep{xu2021audio,liu2018local} \\
    \hline
    \end{tabularx}
    \caption{Summary of encryption techniques used for adding templates not categorized as main template categories.}
    \label{tab:summary_other_sec3}}
\end{table*}

\Paragraph{Encryption in Language Models}
Perturbing probability vectors and using sinusoidal signals to embed templates in language models enhances security.
Encryption in language models as performed by~\citet{zhao2023protecting} involves perturbing the probability vector using a sinusoidal signal and a hash function. The Lomb-Scargle periodogram is employed for spectrum estimation with each text input. The evaluation involves estimating the PSNR using the peak value of the estimated power spectrum at the particular frequency. The process of Poison-only Backdoor Attacks is also adopted for LLM defense by~\citep{li2023black, li2020open, adi2018turning} using templates as triggers on the input images. This attack method subtly corrupts a system, remaining dormant until a specific trigger is activated. The learning objective is tailored to detect such backdoor attacks, using hypothesis testing~\citep{li2023black}, spatial temporal score~\citep{li2020open}, hash functions~\citep{adi2018turning} and similar statistical methods to ascertain the presence of the attack. 
Other methods~\citep{sablayrolles2020radioactive}  perturb the training data with class-dependent isotropic unit vectors to fine-tune the model, making it embed the template into the generated media.

\Paragraph{Gaussian Noise}
Gaussian noise is added to feature points or vertex coordinates in $3$D models to embed templates~\citep{praun1999robust, zhu2021gaussian}. This process adjusts vertex coordinates in key regions or applies DCT transformations to high-frequency coefficients. Robustness and quality are measured through metrics like PSNR, SSIM, and LPIPS. Additionally, random noise templates protect sensitive information while maintaining model accuracy, with differential privacy metrics like $\epsilon$ used to assess privacy levels~\citep{dwork2006differential, zhang2018privacy, cohen2019certified}.

\Paragraph{Trigger Samples and Predefined Tags}
%Using trigger samples and predefined tags enhances model robustness and detection capabilities.
Many works use trigger samples added to the input samples~\citep{zhang2023black, lim2022protect, kapusta2023protecting, ahmadi2020redmark}, and the neural networks are then trained/finetuned with these trigger-containing samples. The extraction process involves calculating correlations between extracted trigger samples and the original trigger pattern or by estimating the outputs, which would only be predicted if trigger samples are used. Success rates of trigger verification and model accuracy on trigger samples are used to assess effectiveness. Metrics such as accuracy, precision, recall, F1 score, and trigger recognition rates are also employed. Predefined tags, such as AprilTags~\citep{wang2016apriltag, olson2011apriltag, garrido2014automatic}, are embedded in images or $3$D models for applications like pose estimation and automatic detection. The learning process involves designing fiducial markers and training detection algorithms to recognize and decode these tags. The extraction process involves detecting and decoding these tags using specialized algorithms. Metrics include detection range, robustness to occlusion, and computational efficiency.

\Paragraph{Code Protection and Obfuscation}
Reverse engineering and obfuscation techniques safeguard code integrity and prevent unauthorized access. In code protection, reverse processes convert bit sequences back into operands and opcodes, revealing templates through dummy methods to verify authenticity and detect tampering~\citep{monden2000practical}.~\citet{balachandran2014function} obfuscate code by transforming instruction sequences ending with jump instructions into basic blocks, displacing them to different functions. The obfuscation's effectiveness is assessed using tools like IDA Pro, focusing on disassembly and control flow errors to ensure robustness against automated attacks.

\Paragraph{Miscellaneous}
Fourier space encryption, passport layers, frontier stitching, and privacy preservation techniques offer robust methods for enhancing security and privacy in digital content. Fourier space encryption, as demonstrated by~\citet{wen2023tree}, embeds templates into the Fourier space of a noise array using a rotation-invariant pattern composed of concentric rings selected from a Gaussian distribution. This method provides resilience to various image transformations and evaluates template presence by estimating a P-value, with detection confirmed if the P-value falls below a predetermined threshold, $\alpha$. Passport layers, introduced by~\citet{fan2019rethinking}, enhance model security by adjusting the scale factor and bias terms of convolutional layers, ensuring that the network functions correctly only with the proper `passport' parameters. An incorrect passport distorts the output, preventing unauthorized use and reverse engineering, while the non-invertibility of the design further secures it against tampering.

Frontier stitching and privacy preservation methods further contribute to the security landscape. The frontier stitching algorithm by~\citet{le2020adversarial} subtly marks a model by clamping its decision frontier, using hypothesis testing to verify the template's presence.~\citet{laarhoven2019nearest} employ hash tables to manage high-dimensional data, identifying near neighbors and detecting potential colluders by analyzing data through sparse dot products, useful in large datasets. For privacy preservation,~\citet{xu2021audio} propose incorporating audio as noise by extracting it from a video, mapping it into a low-dimensional space, and adding it to the video frames' codebook. This ensures that only authorized receivers can decode and reconstruct the original video frames. Additionally,~\citet{liu2018local} use a spatiotemporal chaotic system for chaotic cryptography, applying an improved encryption method based on a spatial chaotic map to secure the face region within an image, enhancing privacy and resisting decryption efforts.

In conclusion, these diverse approaches extend beyond conventional methods to embed and extract templates, enhancing the security and robustness of digital content. Techniques such as Fourier space encryption, Gaussian noise in $3$D models, trigger samples, and predefined tags offer innovative solutions for various applications. The integration of advanced encryption, obfuscation, and privacy-preserving methods ensures the integrity and protection of data across multiple applications. These advancements underscore the importance of developing versatile and resilient techniques to safeguard digital information in an increasingly complex and interconnected world.

\section{Tasks}
Now that we have a good understanding for the types of templates and different process involved like encryption and learning process. Now we provide a discussion on the application of these proactive techniques in various applications. These techniques are used in a variety of applications, which are summarized in~\cref{tab:summary_of_application}. 

\subsection{Vision Models Defense}
In the evolving landscape of digital media, vision models defense has become a critical area of focus to ensure the integrity, authenticity, and security of visual content. This field encompasses a range of techniques to detect, prevent, and attribute deepfakes and other forms of tampered media.

\Paragraph{Deepfake Detection and Attribution}
Enhancing deepfake detection and source attribution involves techniques like artificial fingerprints and learnable templates.~\citet{yu2021artificial} use artificial fingerprints from training data to identify and attribute deepfakes to their source models.~\citet{Asnani_2022_CVPR, Asnani_2023_CVPR} introduce learnable templates in real images for improved detection and localization of tampered images. The DeepMark~\citep{tang2024deepmark} framework uses a Digital Metadata Marker (DMM) for scalable deepfake detection by comparing visual features. AdvMark~\citep{wu2024watermarks} embeds adversarial templates as templates to enhance deepfake detection accuracy.~\citet{asnnai_reverse} estimates the fingerprints left by generative models using the predefined constrains for deepfake detection, image attribution and reverse engineering of generative models.
\begin{figure}[t!]
\centering
\includegraphics[trim={0 -4 0 0},clip,width=\columnwidth]{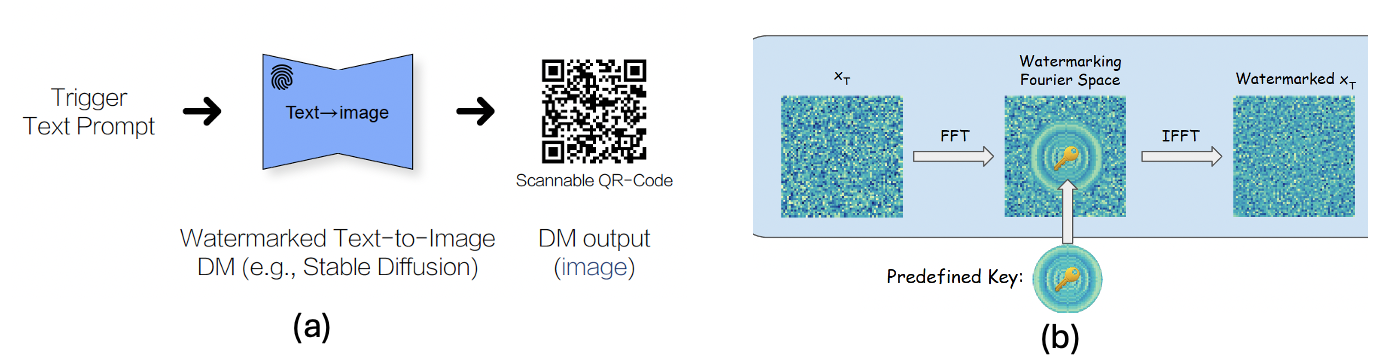}
\caption{Adding tags and triggers to the images. (a) Embedding trigger text prompt to the model which when given the prompt, would output the QR code!\citep{zhao2023recipe}, and (b) adding a predefined tag to the Fourier space of the image~\citep{wen2023tree}.}
\label{fig:sec_3_tag_method}
\vspace{-4mm}
\end{figure}

\Paragraph{Tampering Detection and Verification}
Watermarking and embedding techniques ensure image integrity and recovery after tampering. Various methods, such as spatial domain embedding~\citep{laouamer2015robust}, DCT-based schemes~\citep{singh2016effective}, and singular value decomposition~\citep{dadkhah2014effective}, create templates to detect and sometimes recover tampered content. These techniques maintain image integrity by detecting altered blocks and recovering areas with high precision. Adaptive strategies further enhance this by considering image block complexity and employing techniques like image smoothness differentiation, overlapping embedding~\citep{hsu2016image}, and hierarchical recovery~\citep{qin2017fragile, cao2017hierarchical, hsu2010probability}. Interlocking templates within image blocks~\citep{lee2008dual, haghighi2018trlh} enhance tamper detection and provide fallback for recovery, improving digital media resilience.

\Paragraph{Face Anti-Spoof}
~\citet{yu2023visual} proposes to use proactive methods for face anti-spoofing. The work addresses the challenge of missing modalities in both training and testing phases by incorporating visual prompts and residual contextual prompts in multimodal transformers, ensuring robust learning of flexible-modal features with minimal computational overhead.

\Paragraph{Identity Protection} Embedding authentic signatures in images and using dual traces help protect personal identities from deepfakes. Techniques like face feature disentanglement combined with encryption embed authentic signatures into digital images, ensuring identity protection~\citep{zhao2023proactive}. Dual traces, both sustainable and erasable, verify authenticity and detect fraud in media~\citep{sun2023faketracer, zhang2023editguard}. Additionally, encoder-decoder methods, enhanced by differentiable JPEG compression~\citep{yang2021faceguard}, defend against deepfakes by detecting compression artifacts. 

\begin{figure}[t!]
\centering
\includegraphics[trim={0 -4 0 0},clip,width=\columnwidth]{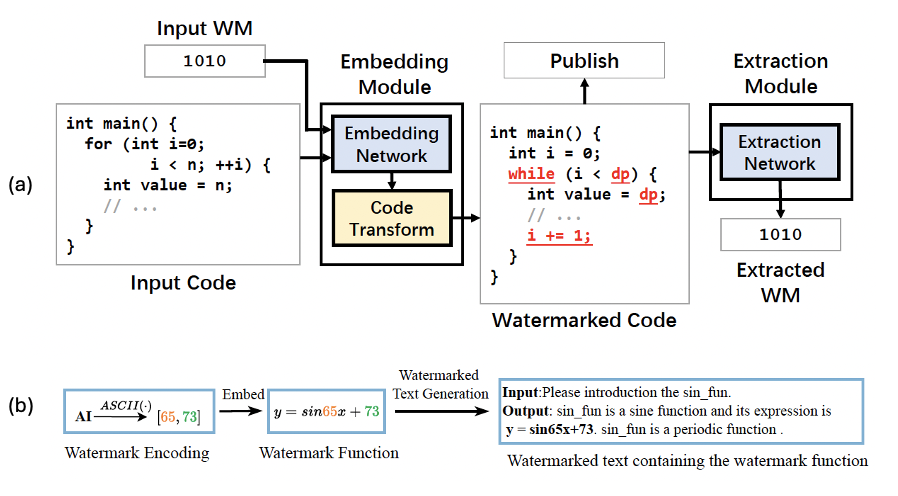}
\caption{Method using other type of templates like (a) code transformations~\citep{yang2023towards}, and (b) specific functions~\citep{li2023functionmarker} into the data to embed templates into the input data.}
\label{fig:sec_3_other_method}
\vspace{-4mm}
\end{figure}

\Paragraph{Disrupting Deepfake Generation}
To protect individual identities, the encoder-decoder approach trained with generative models preserves identity nuances~\citep{wang2021faketagger}. Introducing targeted noise as templates preempts deepfake generation, ensuring models trained on perturbed images produce subpar results~\citep{van2023anti, huang2022cmua, segalis2020ogan, ruiz2020disrupting, yeh2020disrupting}. This proactive defense undermines deepfake quality, making them less convincing and more detectable. Techniques by~\citep{segalis2020ogan, lu2001multipurpose} create images resistant to manipulation, protecting against face-swapping attempts. Adversarial attacks on image translation networks~\citep{yeh2020disrupting} compromise deepfake generation, resulting in flawed outputs. The DUAW technique~\citep{ye2023duaw} disrupts the variational autoencoder (VAE) in Stable Diffusion models, introducing distortions to protect images universally.

%Together, these diverse approaches create a comprehensive defense strategy against deepfakes. Techniques like deepfake detection and attribution, tampering detection, face feature disentanglement, and disrupting deepfake generation all contribute to enhancing the security and authenticity of digital media. These advancements reflect a strong commitment to improving digital security, preserving content integrity, and maintaining the trustworthiness of media in the digital age.

\begin{table*}[hbt!]
{\footnotesize % Reduce font size for the table
    \centering
    \begin{tabularx}{\textwidth}{|>{\centering\arraybackslash}m{3.1cm}|>{\centering\arraybackslash}m{5.2cm}|>{\centering\arraybackslash}m{1.05cm}|>{ \tiny\centering\arraybackslash}m{6.5cm}|}
    \hline
    \rowcolor[HTML]{EFEFEF}  \textbf{Application} & \textbf{Description} & \textbf{Type of Template} & \textbf{References} \\
    \hline
     Deepfake Detection and Attribution & Techniques for identifying and attributing deepfake content. & Bit seq., Learn per. &~\citep{yu2021artificial, Asnani_2022_CVPR, Asnani_2023_CVPR, tang2024deepmark, wu2024watermarks} \\
    \hline
     Tampering detection and verification & Techniques to detect tampering and ensure image integrity. & Bit seq., $2$D noise &~\citep{laouamer2015robust, singh2016effective, dadkhah2014effective, hsu2016image, qin2017fragile, cao2017hierarchical, hsu2010probability, lee2008dual, haghighi2018trlh} \\
    \hline
     Face Anti-Spoof & Detecting face spoofing. & V.P. &~\citep{yu2023visual} \\
    \hline
     Identity Protection & Protect personal identities against deepfakes. & Learn per., Bit seq. &~\citep{zhao2023proactive, sun2023faketracer, zhang2023editguard, neekhara2022facesigns, yang2021faceguard} \\
    \hline
     Disrupting deepfake generation & Disrupt deepfake generation and ensure image resistance. & Per. &~\citep{wang2021faketagger, van2023anti, huang2022cmua, segalis2020ogan, ruiz2020disrupting, yeh2020disrupting, ye2023duaw} \\
    \hline
     Techniques for authenticity verification and provenance tracking & Authenticity verification and provenance tracking. & Bit seq., Text &~\citep{meng2022traceable, zhao2023protecting, wu2023sepmark, munyer2023deeptextmark} \\
    \hline
     Defensive Strategies Against Malicious Exploitation & Enhance LLM resilience against malicious exploitation. & Text, triggers &~\citep{robey2023smoothllm, dong2023revisit, liu2023watermarking} \\
    \hline
     Context-Aware Modifications & Encryption via probabilistic outputs and context-aware lexical substitutions. & Text &~\citep{kirchenbauer2023reliability, yang2022tracing, he2022protecting, rizzo2019fine, kirchenbauer2023watermark} \\
    \hline
     Intellectual Property Protection & Methods to protect intellectual property. & Text &~\citep{zhang2010reference, he2022cater, yang2023towards, li2023functionmarker} \\
    \hline
     Model Attribution & Content origin identification. & Bit seq. &~\citep{zeng2023securing, wen2023tree, wu2020watermarking, zhao2023recipe, atli2022effectiveness, yu2019attributing} \\
    \hline
     Neural Network Ownership and Protection & Embed templates within neural networks to protect and prove ownership of models. & V.P., per., bit seq. &~\citep{darvish2019deepsigns, xue2022advparams, chen2019deepmarks, adi2018turning, uchida2017embedding, nagai2018digital, le2020adversarial, fernandez2023stable, liu2021watermarking, zhang2018protecting, peng2022fingerprinting} \\
    \hline
     Ownership Verification in Federated Learning & Ownership verification of federated neural networks. & Bit seq., per. &~\citep{li2022fedipr, han2022verifiable, tekgul2021waffle, xu2019verifynet, liu2021secure} \\
    \hline
     Protection Techniques for Diffusion Models & Protect diffusion models. & Bit seq., per. &~\citep{yang2024gaussian, huang2024freezeasguard, cui2023ft, lei2024diffusetrace, meng2024latent, ahmadi2020redmark, zhang2024robust, zhang2024training, peng2023intellectual, lim2022protect} \\
    \hline
     Camera Model Encryption and Localization & Enhance forensic analysis and image authenticity verification. & Bit seq., per. &~\citep{cozzolino2019noiseprint, wu2023sepmark} \\
    \hline
     Data and Artists' Attribution & Ensure proper recognition and preservation of authorship rights for data and artists. & Bit seq. &~\citep{cui2023diffusionshield, asnani2024promark, li2023black, li2020open} \\
    \hline
     Fingerprinting for preservation of authorship rights & Identifying authorship rights infringement and protecting digital content. & Bit seq. &~\citep{furon2014tardos, balachandran2014function, fan2019rethinking} \\
    \hline
     User Privacy Preservation & User privacy preservation ensuring efficient protection of personal data. & Per. &~\citep{laarhoven2019nearest, li2021visual, dwork2006differential, zhang2018privacy, blum2005practical, zhang2021proactive, tang2024once, li2023exploring} \\
    \hline
     Face Recognition Privacy & Face recognition protection to deceive models & Per. &~\citep{xiao2021improving, zhong2020towards, rajabi2021practicality, shan2020fawkes, wu2019privacy, hu2022protecting, xu2021audio, komkov2021advhat, dong2019efficient} \\
    \hline
     Miscellaneous & Autonomous driving privacy, and surveillance privacy. & Per., bit seq. &~\citep{mirjalili2018semi, xiong2020adgan, paruchuri2009video, liu2018local} \\
    \hline
     Point Cloud Adversarial Defense and Encryption &Enhancing robustness of $3$D data models and encryption for point clouds. & Per., Bit seq. &~\citep{liang2022pagn, ding2021point, liu2019novel, xiaoqing2015watermarking, feng2015new, ohbuchi2004watermarking, cotting2004robust, ohbuchi2001watermarking} \\
    \hline
     $3$DMMs, SDFs & $3$DMMs encryption ensuring preservation of authorship rights and content authentication. & Bit seq. &~\citep{wang2022neural, zhu2023towards} \\
    \hline
     NeRF Models Encryption & Encryption methods for Neural Radiance Fields (NeRF) for preservation of authorship rights. & Bit seq. &~\citep{luo2023copyrnerf, jang2024waterf, chen2024nerf, chen2023marknerf, li2023steganerf, huang2024noise} \\
    \hline
     $3$D Mesh Defense & Embedding templates in $3$D meshes to ensureauthorship rights and tamper detection. & Bit seq. &~\citep{hamidi2019blind, wang2022deep, zafeiriou2005blind, yoo2022deep, praun1999robust, zhu2024rethinking, zhu2021gaussian, medimegh20183d} \\
    \hline
     $3$D GS Protection & Embedding hidden information in $3$D scenes. & Bit seq. &~\citep{zhang2024gs} \\
    \hline
     $3$D Models Protection & Protection for $3$D models ensuring minimal visibility of distortions. & Bit seq. &~\citep{peng2022semi, benedens1999geometry, nakazawa2010visually, yeung1998fragile, alface2007blind, kanai1998digital, liu2012spectral, liu2012three, chou2006public, chou2009affine} \\
    \hline
     VLMs & Enhancing VLMs for downstream applications. & V.P., T.P. &~\citep{kunananthaseelan2024lavip, zhang2023text, zhu2023visual, wu2022unleashing, nasiriany2024pivot, xing2023dual, shen2024multitask, zhao2023prompting, mirza2024meta, maniparambil2023enhancing} \\
    \hline
     Visual Prompt Tuning for GMs & Enhancing image generation quality and efficient transfer learning. & V.P. &~\citep{zhang2024instruct, sohn2023visual, kim2024we, ma2024learning, han20232vpt, kim2024we, park2024fair, wu2022promptchainer, chen2023understanding, ju2022prompting, wang2024revisiting, zhang2024visual, yoo2023improving, song2023deep} \\
    \hline
     Text-to-3D Generation & Enhancing text-to-3D generation. & V.P. &~\citep{chen2024vp3d} \\
    \hline
     Object Localization and Tracking & Unadversarial examples, fractal markers, object recognition, localization, and tracking. & Bit seq., V.P. &~\citep{salman2021unadversarial, asnani2024probed, wagner2007artoolkitplus, wang2016apriltag, olson2011apriltag, garrido2014automatic, abbas2019analysis, alvarez2012new, wang2018hierarchical} \\
    \hline
     Image Editing and Inpainting & Text-based image editing and inpainting. & V.P. &~\citep{nguyen2024visual, bar2022visual} \\
    \hline
     Medical Image Segmentation & Improve medical image segmentation. & V.P. &~\citep{wang2023pcdal, wang2023fvp} \\
    \hline
    \end{tabularx}
    \caption{Summary of categories, descriptions, types of templates, and references. [KEYS: seq.: sequence, V.P.: visual prompt, per.: perturbation]}
    \label{tab:summary_of_application}}
\end{table*}

\subsection{LLM Defense}

Ensuring the security and integrity of Large Language Models (LLMs) is critical as they face risks like unauthorized use, disinformation, and adversarial attacks. A comprehensive defense strategy now includes authenticity verification, provenance tracking, and robust protections against malicious activities to safeguard LLMs in the digital age.

\Paragraph{Authenticity Verification and Provenance Tracking}
Ensuring authenticity and tracking content provenance are crucial in combating disinformation. Techniques like the Decoupled Invertible Neural Network (DINN)~\citep{meng2022traceable} encode dual-tags into images as fingerprints for authenticity verification and provenance tracking. Advanced text encryption methods interweave encoded signals within natural language~\citep{zhao2023protecting}, using neural networks~\citep{wu2023sepmark} or linguistic tools like the Universal Sentence Encoder and Word2Vec~\citep{munyer2023deeptextmark}. These methods also secure code by embedding templates, protecting against plagiarism and unauthorized use.

\Paragraph{Defensive Strategies Against Malicious Exploitation}
Various defensive strategies have been developed to protect against malicious exploitation. Character-level input perturbations~\citep{robey2023smoothllm} inoculate against adversarial attacks, and specialized datasets~\citep{dong2023revisit} enhance the resilience of language models under real-world conditions. Encryption strategies have evolved to include backdoor techniques that insert covert triggers into text outputs~\citep{liu2023watermarking}, enabling the tracing of unauthorized reproductions.

\Paragraph{Context-Aware Modifications}
Embedding secret signals and making context-aware modifications ensure that templates remain undetected by attackers. Some methods embed secret signals into probabilistic model outputs~\citep{kirchenbauer2023reliability} or apply context-aware lexical substitutions~\citep{yang2022tracing, he2022protecting}, detectable only by insiders. Homoglyph substitutions~\citep{rizzo2019fine} disguise textual input, preserving privacy and message integrity. Additionally, `green tokens'~\citep{kirchenbauer2023watermark} embed templates into high-entropy words to mark and identify the ownership or origin of textual content.

\Paragraph{Restoration-Oriented Approaches and Intellectual Property Protection}
Restoration-oriented approaches and strategies to protect intellectual property are crucial for maintaining content integrity. Self-embedding template schemes~\citep{zhang2010reference} mark and aid in recovering original content if tampered with. To protect text generation APIs, subtle alterations in word distribution patterns create hard-to-detect templates~\citep{he2022cater}. Code transformation and variable substitution generate template functions~\citep{yang2023towards, li2023functionmarker}, which can be verified later for user code protection.

%In conclusion, diverse strategies such as dual-tag encoding, text encryption, and adversarial defenses highlight the importance of innovation in securing language models. These methods strengthen model resilience and protect intellectual property, ensuring ethical use and reducing misuse. As technology advances, ongoing commitment to these security measures will be vital for maximizing the potential of language models while managing associated risks.

\subsection{Attribution and preservation of authorship rights}
The applications and methods summarized in this section illustrate a broad range of innovative strategies employed for securing generative models, tracing sources, and establishing neural network ownership and preservation of authorship rights through various forms of proactive schemes and fingerprinting.

\Paragraph{Model Attribution}
Proactive techniques for attributing model outputs help ensure content origin identification. One approach involves fine-tuning models with encrypted images to perform effective attribution. Techniques by~\citet{zeng2023securing},~\citet{wen2023tree}, and~\citet{wu2020watermarking} use unique characteristics of encrypted data to trace and secure generative models (GMs), embedding identifiable signals within the data. Additionally, encoding binary strings into diffusion models~\citep{zhao2023recipe, atli2022effectiveness, yu2019attributing} involves fine-tuning with encrypted image pairs and trigger prompts to trace content back to its rightful owner.

\Paragraph{Neural Network Ownership and Protection}
The application of neural network protection and ownership is addressed through various methods to embed templates within networks. Techniques like DeepSigns~\citep{darvish2019deepsigns, xue2022advparams} embed information into the probability density function of activation sets, while DeepMarks~\citep{chen2019deepmarks} offers an end-to-end fingerprinting framework. These methods resist attacks like model extraction and collusion, securing creative ownership.
Other approaches include backdoor attacks~\citep{adi2018turning}, additional regularizers~\citep{uchida2017embedding, nagai2018digital, le2020adversarial}, fine-tuning the latent decoder of diffusion models~\citep{fernandez2023stable}, network parameters residuals~\citep{liu2021watermarking}, query prompts~\citep{zhang2018protecting}, and learnable perturbations~\citep{peng2022fingerprinting}. A deep spatial encryption framework~\citep{zhang2021deep} is robust against surrogate model attacks, supporting image-based templates to protect data and algorithms.
Another method involves embedding a template into the weights of a neural network~\citep{wang2020watermarking}, ensuring robustness against brute-force attacks. The DAWN framework~\citep{szyller2021dawn} deters model extraction by dynamically changing responses to specific queries. BlackMarks~\citep{chen2019blackmarks} ensures fidelity, robustness, and security for intellectual property protection. %RIGA~\citep{wang2021riga} embeds templates without impacting model accuracy. Other methods~\citep{wang2023plug, kapusta2023protecting, namba2019robust, guo2018watermarking} advance research on protecting DNNs against malicious purposes.

\begin{figure}[t!]
\centering
\includegraphics[trim={0 -4 0 0},clip,width=\columnwidth]{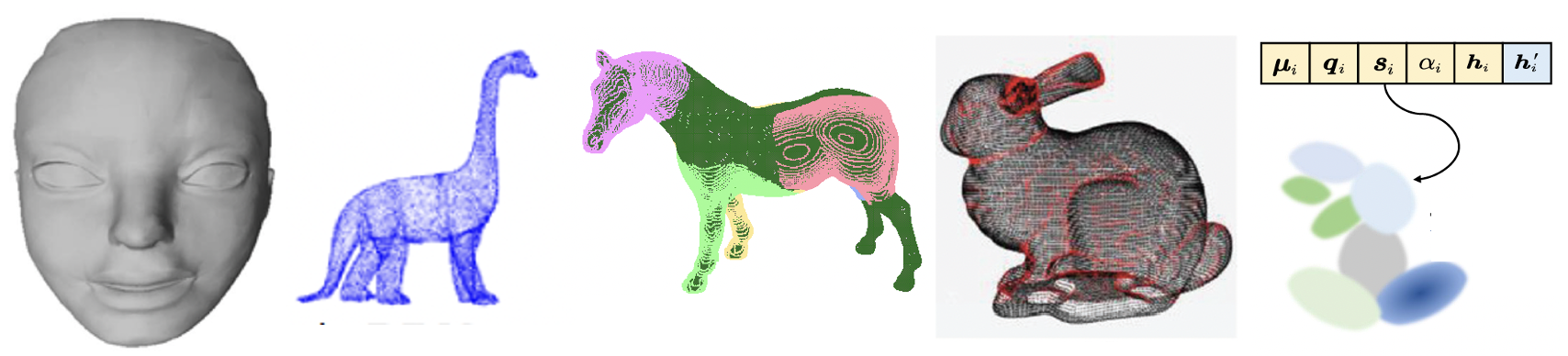}
\caption{Template addition in $3$D meshes~\citep{yu2003robust, liu2019novel, zhang2024gs},  point clouds meshes~\citep{feng2015new}, point cloud, and into the $3$D vertices~\citep{ohbuchi2002frequency}. }
\label{fig:3d_domain_example}
\vspace{-4mm}
\end{figure}

\Paragraph{Ownership Verification in Federated Learning}
Ownership verification techniques in federated learning ensure robust and private model ownership. The FedIPR framework~\citep{li2022fedipr} embeds and verifies private templates in Federated Deep Neural Networks (FedDNN) without disclosing information, addressing robustness challenges like client selection and differential privacy.~\citet{han2022verifiable} use key exchange technology and a double masking protocol for privacy protection and correctness verification. WAFFLE~\citep{tekgul2021waffle} embeds resilient templates in DNN models trained with federated learning without accessing training data. %Methods by~\citet{xu2019verifynet},~\citet{liu2021secure}, and~\citet{han2022verifiable} focus on verifiable and privacy-preserving federated learning using key exchange and tag aggregation methods.

\Paragraph{Protection Techniques for Diffusion Models}
Various innovative techniques protect diffusion models while maintaining performance. Gaussian Shading~\citep{yang2024gaussian} embeds templates into generated images using template diffusion, randomization, and distribution-preserving sampling. FreezeAsGuard~\citep{huang2024freezeasguard} selectively freezes critical tensors to prevent unauthorized fine-tuning of diffusion models. FT-Shield~\citep{cui2023ft} uses bi-level optimization and a Mixture of Experts framework to generate and detect templates in text-to-image diffusion models. DiffuseTrace~\citep{lei2024diffusetrace} embeds invisible templates into generated images without compromising quality, while Latent template~\citep{meng2024latent} injects and detects templates in the latent space of diffusion models. %Additional methods~\citep{ahmadi2020redmark, zhang2024robust, zhang2024training, peng2023intellectual, lim2022protect} further enhance protection capabilities.

\Paragraph{Camera Model Encryption and Localization}
Camera model encryption techniques enhance forensic analysis and image authenticity verification.~\citet{cozzolino2019noiseprint} use a Siamese network trained with image patches from different cameras to detect image forgeries and identify the specific camera model, aiding forensic analysis and ensuring digital image credibility. SepMark~\citep{wu2023sepmark} embeds an encoded message into pristine images, allowing for later extraction to verify authenticity and trace the image source, providing robust deepfake defense and source tracing capabilities.

\Paragraph{Data and Artists' Attribution}
Techniques for data and artist attribution ensure proper recognition and authorship rights. ProMark~\citep{asnani2024promark} encrypts training data with bit sequence templates, enabling GenAI models to perform concept attribution during training. Diffusion Shield~\citep{cui2023diffusionshield} embeds unique messages into datasets for identification without retraining. Poison-only backdoor attacks~\citep{li2023black, li2020open} mark datasets by embedding special behaviors in data samples, allowing verification of model training origins through hypothesis testing.

\Paragraph{Fingerprinting for preservation of authorship rights}
Active fingerprinting techniques provide powerful tools for identifying authorship rights infringement and protecting digital content. These techniques include the use of Tardos codes for traitor tracing~\citep{furon2014tardos}, code fragments relocation~\citep{balachandran2014function}, and passport-based DNN ownership verification schemes~\citep{fan2019rethinking}. By embedding digital passports and employing active fingerprinting, content creators can more effectively control and enforce their authorship rights in the digital realm, preventing unauthorized use and distribution.

%Each of the above works demonstrates a commitment to enhancing the security, integrity, and ownership rights of digital content and models through embedding signals as templates. By addressing the challenges associated with model attribution, source tracing, and preservation of authorship rights, these methods offer valuable contributions to the field of digital security and intellectual property management.

\subsection{Privacy Protection}
The collection of works discussed in this section showcases a concerted effort to preserve privacy across various digital platforms, with a particular focus on the protection of user identity and personal data in the face of advanced facial recognition technologies and surveillance mechanisms.

\Paragraph{User Privacy Preservation}
Methods aimed at user privacy preservation ensure efficient protection of personal data. The score-based fingerprinting framework~\citep{laarhoven2019nearest} accelerates decoding times for efficient data protection. Mapping Distortion Based Protection (MDP) and AugMDP~\citep{li2021visual} misalign images and labels to confuse potential data breaches without compromising benign neural network performance.
Various techniques protect privacy on digital platforms, including adding random noise to query functions for differential privacy~\citep{dwork2006differential}, obfuscation techniques for training data~\citep{zhang2018privacy}, and the SuLQ framework for statistical database privacy~\citep{blum2005practical}. Proactive Privacy-preserving Learning (PPL)~\citep{zhang2021proactive} uses adversarial generators to transform data for malicious model manipulation. Universal Transferable Adversarial Perturbation (UTAP)\citep{tang2024once} protects privacy in facial image databases. %Prom-PATE~\citep{li2023exploring} combines visual prompting with Private Aggregation of Teacher Ensembles (PATE) to enhance privacy-utility trade-offs in deep learning. ADAF~\citep{wu2023towards} defends facial privacy in text-to-image diffusion models.

\Paragraph{Face Recognition Privacy}
Techniques for face recognition protection deceive facial recognition models without affecting legitimate applications. Methods like adversarial patches and perturbations are finely tuned to deceive models~\citep{xiao2021improving, zhong2020towards, rajabi2021practicality}, while systems like Fawkes~\citep{shan2020fawkes} apply pixel-level changes to prevent unauthorized recognition. Other approaches perform deidentification in the feature space~\citep{wu2019privacy}.
Innovative adversarial and encryption techniques include ATM-GAN~\citep{hu2022protecting}, which creates adversarial examples to distort makeup styles and prevent unauthorized recognition. The cycle-VQ-VAE framework obscures video streams by integrating audio as noise~\citep{xu2021audio}.~\citet{komkov2021advhat} use a Spatial Transformer Layer to project stickers onto face images for face ID systems.~\citet{dong2019efficient} discuss black-box adversarial attacks on face recognition systems, exposing deep CNN vulnerabilities. %AnonymousNet~\citep{li2019anonymousnet} combines facial attribute estimation, face obfuscation, image synthesis, and adversarial perturbation for privacy and security. Another method uses Delaunay triangulation and affine transformations to protect privacy while retaining biometric utility~\citep{mirjalili2017soft}.

\Paragraph{Autonomous Driving and Surveillance}
In autonomous driving, ADGAN~\citep{xiong2020adgan} protects location privacy by obscuring sensitive information in camera data while maintaining utility. Similarly, surveillance privacy techniques, such as those by~\citet{paruchuri2009video} and~\citet{liu2018local}, use spatial chaotic maps to encrypt human faces in video footage, safeguarding identities without compromising the overall utility of the footage.

%Collectively, these works represent a robust defense against the misuse of personal data and unauthorized identity recognition, employing a blend of adversarial attacks, data distortion, selective attribute protection, and encryption to tackle the ever-growing challenges of digital privacy.

\subsection{3D Domain}

\begin{figure}[t!]
\centering
\includegraphics[trim={0 -4 0 0},clip,width=\columnwidth]{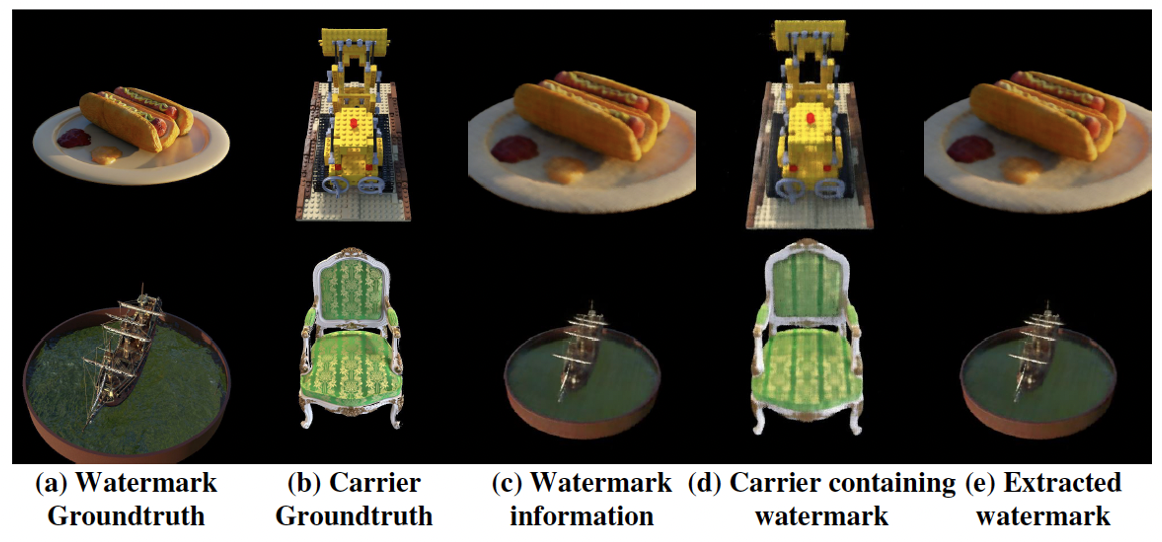}
\caption{Input-encrypted input pairs for NeRF models~\citep{luo2023copyrnerf, chen2024nerf}. }
\label{fig:nerf_example}
\vspace{-3mm}
\end{figure}

In $3$D modeling, protecting digital content and intellectual property is crucial across industries like entertainment, manufacturing, medical imaging, and virtual reality. Proactive methods, including encryption techniques for $3$D models, meshes, and point clouds, embed imperceptible yet resilient templates into digital assets. These techniques detect unauthorized use, prevent tampering, and ensure preservation of authorship rights and content authentication. %Below, we discuss proactive methods used for various applications in the $3$D domain.

\Paragraph{Point Cloud Adversarial Defense and Encryption}
%Proactive techniques in point cloud adversarial defense enhance $3$D data processing model robustness.
The Perturbation Adaption Generation Network (PAGN)~\citep{liang2022pagn} is designed for point cloud adversarial defense in $3$D model classification. PAGN includes a perturbation-injection module, a generative module, and a shape similarity measure to improve robustness. The perturbation-injection module simulates adversarial samples, while the generative module visualizes these samples and measures shape similarity. Additionally,~\citet{ding2021point} propose the Geometry-Consistent Point Cloud Upsampling (GC-PCU) method, generating uniform, clean, and dense point clouds from sparse ones through feature extraction, perturbation learning, and geometric reconstruction.
Encryption techniques for $3$D point clouds balance transparency and robustness~\citep{liu2019novel, xiaoqing2015watermarking, feng2015new, ohbuchi2004watermarking, cotting2004robust}. Methods include defining local sets, calculating RMSC values, establishing synchronization relations, and embedding templates by modifying distance normalization means~\citep{liu2019novel}. %Principal component analysis (PCA) aligns models post-transformations, maintaining template effectiveness despite affine transformations~\citep{xiaoqing2015watermarking}. For $3$D polygonal meshes, encryption involves embedding data into vertex coordinates or topology~\citep{ohbuchi2001watermarking, ohbuchi1998watermarking} using methods like TSQ, TVR, and Mesh Density Pattern algorithms. These methods ensure robustness against mesh simplification, remeshing, and noise addition by embedding in spectral coefficients~\citep{ohbuchi2002frequency}, vertex norms~\citep{cho2006oblivious, kuo2009blind}, or geometric modifications~\citep{benedens2000towards}.

\Paragraph{3D Morphable Models and Signed Distance Fields Encryption}
Encryption techniques for $3$D Morphable Models (3DMMs) ensure preservation of authorship rights and content authentication. A deep neural network scheme~\citep{wang2022neural} uses an encoder to encode templates and meshes, an Attacker to add perturbations, and a Decoder to extract templates. The FuncMark method~\citep{zhu2023towards} embeds binary templates in signed distance fields (SDFs) using spherical partitioning and local deformation, allowing template extraction from derived meshes.

\Paragraph{NeRF Model Encryption}
Neural Radiance Fields (NeRF) have advanced preservation of authorship rights through neural $3$D encryption. One method trains a $2$D decoder and the NeRF model separately, achieving high bit accuracy and reconstruction quality using patch loss and discrete wavelet transform~\citep{jang2024waterf}. Another approach uses Implicit Neural Representation (INR) to embed template information into NeRF models, addressing low template capacity and security risks~\citep{chen2024nerf}. MarkNeRF~\citep{chen2023marknerf} uses neural networks to protect implicit data representations, offering high imperceptibility, robustness, and anti-interference capability. StegaNeRF~\citep{li2023steganerf} embeds invisible information within NeRF renderings through a two-stage optimization process, balancing rendering quality and decoding accuracy. Noise-NeRF~\citep{huang2024noise} introduces trainable noise on specific views for steganography, enhancing quality and efficiency.

\Paragraph{3D Mesh Defense}
%Techniques for $3$D mesh defense ensure preservation of authorship rights and detect tampering. 
$3$D mesh defense techniques involve embedding templates as templates into the geometrical structures of $3$D models to ensure preservation of authorship rights and detect tampering~\citep{hamidi2019blind, wang2022deep, zafeiriou2005blind, yoo2022deep, praun1999robust}. Methods such as DEEP3DMARK~\citep{wang2022deep} use attention-based convolutions to embed templates into $3$D meshes. Another approach uses wavelet transform~\citep{hamidi2017robust, wang2008hierarchical} and the norm of wavelet coefficient vectors as encryption primitives, embedding templates by quantizing these norms~\citep{wang2011robust}. Some techniques employ Voronoi patches~\citep{ai2009new} and transform applications for template embedding, while others use robust mesh feature segmentation~\citep{feng2014double} and DCT transformation. Many of these methods are designed to be resistant to various attacks, including noise addition, $3$D rotation, simplification, and cropping.

\Paragraph{3D Gaussian Splatting Protection}
%Techniques for $3$D Gaussian splatting (3DGS) protection enhance security and transparency in $3$D scenes. 
The GS-Hider framework~\citep{zhang2024gs} embeds hidden information securely using a coupled secured feature attribute and rendering pipeline. It employs two parallel decoders to separate rendered RGB scenes and hidden messages, ensuring robustness against rendering. This method enhances security, transparency, and authenticity in encrypted transmission, $3$D compression, and preservation of authorship rights.

\Paragraph{3D Model Protection}
Protection techniques for $3$D models ensure minimal visibility of distortions while maintaining security. Methods involve modifying the geometrical structure or vertex positions to embed bits according to a key~\citep{peng2022semi, benedens1999geometry, nakazawa2010visually, yeung1998fragile, alface2007blind, kanai1998digital}. Templates are used for data hiding to protect models while maintaining data integrity~\citep{hou2023separable, jiang2017reversible, zhang2023adaptive, tsai2022integrating}. Techniques like octree spatial subdivision and multi-MSB prediction enhance embedding capacity and lossless recovery~\citep{hou2023separable}.~\citet{jiang2017reversible} use coordinate transformation, prediction error detection, model encryption, and label map embedding to improve embedding rate and capacity.

%Collectively, these works represent a robust defense against the misuse of $3$D digital content and unauthorized intellectual property access. They employ a blend of adversarial defenses, encryption techniques, and encryption to protect and authenticate $3$D models, point clouds, and meshes, ensuring the integrity and security of digital assets in the ever-growing $3$D technology landscape.

\subsection{Improving Generative Models}
We outline various methods for improving the interpretability and generalization of deep neural networks, visual prompt tuning in generative models, transformers, and text-to-3D generation, as well as enhancing vision-language models (VLMs) and large language models (LLMs).

\Paragraph{Vision-Language Models (VLMs)}
Proactive techniques enhance VLMs for downstream applications like image classification, recognition, semantic segmentation, and object detection. Language-Grounded Visual Prompting (LaViP)~\citep{kunananthaseelan2024lavip} uses input-specific visual prompts with language integration for better model adaptability. The Text-to-Video Prompting framework (TVP)~\citep{zhang2023text} improves Text-to-Video Generation (TVG) models with optimized prompts for enhanced generation quality and temporal localization accuracy. Visual Prompt multi-modal Tracking (ViPT)~\citep{zhu2023visual} adapts pre-trained RGB-based models for downstream applications using modality-specific visual prompts. Enhanced Visual Prompting (EVP)~\citep{wu2022unleashing} and Prompting with Iterative Visual Optimization (PIVOT)~\citep{nasiriany2024pivot} enable models to handle spatial applications and multimodal learning without task-specific fine-tuning.
Next, further enhancements in VLMs include techniques like Dual Modality Prompt Tuning (DPT)\citep{xing2023dual} for learning visual and text prompts simultaneously, and MVLPT\citep{shen2024multitask} for incorporating cross-task knowledge into prompt tuning. 
%Methods like DFER-CLIP~\citep{zhao2023prompting} for dynamic facial expression recognition, and Meta-Prompting for Visual Recognition (MPVR)\citep{mirza2024meta} automate prompt generation for zero-shot recognition applications. In improving CLIP models,~\citet{maniparambil2023enhancing} propose incorporating visually descriptive textual (VDT) information generated by GPT-4 into prompts to enhance zero-shot transfer accuracy on specialized fine-grained datasets.

\begin{figure}[t!]
\centering
\includegraphics[trim={0 -4 0 0},clip,width=\columnwidth]{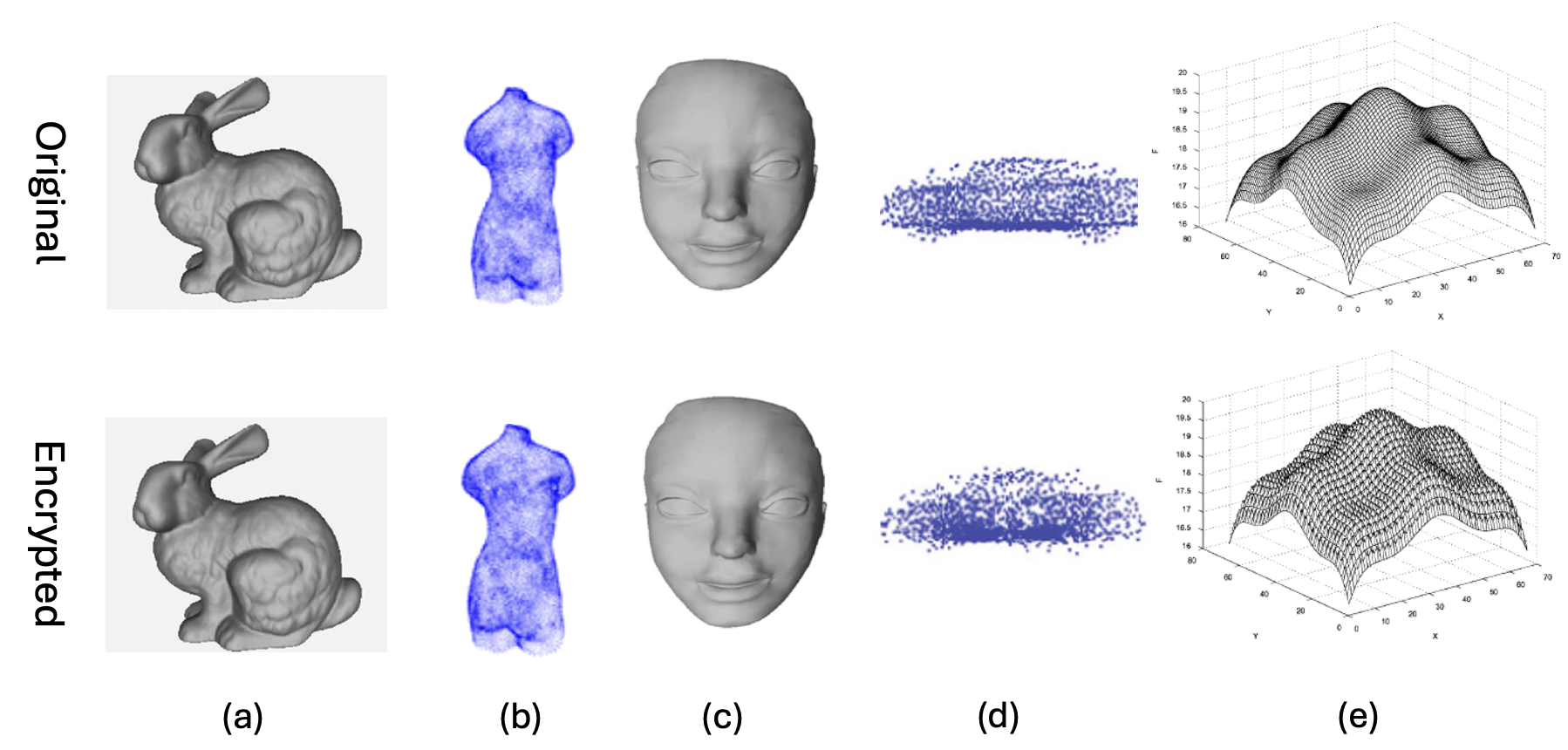}
\caption{Various examples of input-encrypted input pairs after adding the templates into the $3$D input data (a)~\citep{liu2019novel}, (b)~\citep{feng2015new},(c)~\citep{yu2003robust}, (e)~\citep{liang2022pagn}, and (f)~\citep{bors2006watermarking}.}
\label{fig:3d_domain_input_enc_input}
\vspace{-3mm}
\end{figure}

\Paragraph{Visual Prompt Tuning for Generative Models}
In the context of visual prompt tuning for generative models, techniques like InMeMo~\citep{zhang2024instruct} and prompt tuning for generative vision transformers~\citep{sohn2023visual, kim2024we, ma2024learning} focus on enhancing image generation quality and efficient transfer learning. E2VPT~\citep{han20232vpt}, VPT~\citep{kim2024we}, Fair-VPT~\citep{park2024fair}, and PromptChainer~\citep{wu2022promptchainer} aim to improve transformer models by incorporating learnable visual prompts and addressing fairness issues, respectively.
Additional techniques include Iterative Label Mapping-based Visual Prompting (ILM-VP)~\citep{chen2023understanding} for reprogramming pre-trained source models to new target applications, DINOv for generic and referring segmentation applications, and video-based visual-language pre-training (I-VL)~\citep{ju2022prompting} for video understanding applications like action detection and localization, text-video retrieval, summarization, etc. Approaches like Self-Prompt Tuning (SPT)~\citep{wang2024revisiting}, Adaptive Pretraining (AP)~\citep{zhang2024visual}, and Gated Prompt Tuning~\citep{yoo2023improving} adapt pre-trained models to downstream applications. Other methods~\citep{song2023deep, lee2023multimodal, cai2024vip, liu2023explicit} incorporate similar techniques by leveraging prompt tuning to adapt pre-trained models for downstream CV applications.

\Paragraph{Text-to-3D Generation}
For text-to-3D generation, VP3D~\citep{chen2024vp3d} introduces a novel visual prompt-guided diffusion model to enhance text-to-3D generation by utilizing high-quality images from $2$D diffusion models as visual prompts.

%Advancements in proactive schemes, including interpretability, generalization, and prompt tuning, enhance deep neural networks and VLMs. Techniques like visual prompt tuning, causal reasoning, and adaptive pretraining improve performance, fairness, and robustness in tasks like image classification, video understanding, and text-to-3D generation, helping researchers address modern AI challenges with more adaptable models.

\subsection{Other CV applications}

In computer vision, innovative methods are continuously being developed to improve object recognition, localization, tracking, and image manipulation. These advancements are critical for improving the performance and adaptability of models across various applications, from augmented reality to medical imaging.

\Paragraph{Object Localization and Tracking}
To improve object recognition,~\citet{salman2021unadversarial} introduce unadversarial examples, which modify objects to enhance performance and robustness. Their method employs gradient-based algorithms to design unadversarial patches and textures, boosting system resilience in diverse environments.~\citet{asnani2024probed} propose a novel template learning paradigm to improve performance in $2$D and camouflaged object detection.
Several advancements have been made in object localization and tracking~\citep{wagner2007artoolkitplus, wang2016apriltag, olson2011apriltag, garrido2014automatic} reduces false positives, increases detection rates, and minimizes computing time for tag detection, making it suitable for computation-limited systems like smartphones. Fractal markers~\citep{romero2019fractal} provide a novel approach to long-range marker pose estimation, offering robustness to occlusion and wider detection ranges. Further, object detection and segmentation have also improved by adding region proposals~\citep{girshick2014rich} or by using prompts for fine-tuning VLMs~\citep{long2023fine, liang2023visual}.

\Paragraph{Image Editing and Inpainting}
For image editing~\citep{nguyen2024visual} and inpainting~\citep{bar2022visual}, a method involving visual prompting leverages example pairs representing ``before" and ``after" edits to learn text-based editing directions, utilizing text-to-image diffusion models.

\Paragraph{Medical Image Segmentation}
The Progressive Classification and Data Augmentation Learning (PCDAL) framework~\citep{wang2023pcdal} utilizes deep learning models for classification and segmentation applications, incorporating data augmentation to increase training set size and prevent overfitting. Fourier Visual Prompting (FVP)\citep{wang2023fvp} addresses domain shift in medical image segmentation by introducing visual prompts in the frequency domain, guiding pre-trained models to perform well in the target applications.

%Overall, there are a wide range of innovative methods for improving object recognition, localization, tracking, and visual prompting, emphasizing the importance of enhancing model performance and adaptability across various computer vision and medical imaging applications. These advancements ensure more robust, accurate, and versatile models capable of addressing diverse challenges in the field of computer vision.

\section{Challenges}
The landscape of digital content protection is complex, involving intricate relationships between templates, encryption processes, and their applications. Each step in developing robust security measures presents unique challenges.

\Paragraph{Deepfake Detection} Defending against deepfakes with binary sequences and positional values requires balancing visual quality, authenticity, and discardability of templates. Challenges include withstanding identity-switching schemes and generalizing to unknown tampering types, while maintaining robustness against image degradation and manual extraction issues~\citep{sun2023faketracer, meng2022traceable, zhang2023editguard}.

\Paragraph{Textual Content Protection} Protecting LLM-generated text requires encryption methods that maintain semantic integrity while being subtle and versatile across different datasets and models. These methods must function without extensive retraining and survive black-box scenarios~\citep{yang2022tracing, liu2023watermarking}. In code protection, preserving naturalness and operational semantics while embedding resilient templates against obfuscator attacks is crucial~\citep{monden2000practical}.

\Paragraph{Neural Network and Ownership Protection} Ensuring robust ownership protection in neural networks demands encryption that resists fine-tuning, pruning, and overwriting. Templates must be deeply integrated to withstand open-sourcing of models and simple code alterations~\citep{darvish2019deepsigns, adi2018turning, fernandez2023stable}, while allowing public verification without losing credibility~\citep{adi2018turning}.

\Paragraph{Adversarial Perturbation and Face Recognition Protection} Developing effective DNN fingerprinting and adversarial perturbations involves embedding templates that do not impact model performance and are resistant to removal and overwriting~\citep{chen2019deepmarks}. In face recognition, adversarial masks must deceive classifiers while keeping images visually natural, balancing security with user convenience~\citep{yang2021towards, rajabi2021practicality}.

%Each of these challenges is pivotal in advancing proactive security measures, requiring solutions that are as dynamic and adaptable as the evolving threats.

\section{Limitations}

Proactive security schemes, while effective, come with significant limitations. These can be broadly categorized into computational demands, robustness against attacks, generalizability, and practical challenges.

\Paragraph{Computational Demands} Techniques like deepfake defense using fixed bit encodings require substantial computational resources, particularly during pre-processing to adapt signals to various content types~\citep{zhao2023proactive}. The need for redundant message insertion to ensure template survival under transformations further escalates computational costs~\citep{wen2023tree}, making these schemes resource-intensive.

\Paragraph{Robustness Against Attacks} The effectiveness of these schemes is often compromised by their susceptibility to adversarial attacks. For example, attackers can retrain models using the same datasets employed for encryption, weakening the security of generative models~\citep{zeng2023securing}. Adversarial noise can disrupt encrypted messages in deepfake defenses~\citep{wang2021faketagger}, and template manipulations can significantly alter computational costs and text quality~\citep{kirchenbauer2023watermark}. Even code protection methods using template embedding in Java files are at risk of being completely negated by additive attacks~\citep{monden2000practical}.

\Paragraph{Generalizability and Semantic Integrity} A major challenge for these schemes is their limited generalizability across different types of content. Binary sequences and Fourier key templates, while effective for specific applications, often cause image distortion and fail to work well with dynamic content like video or 3D scenes~\citep{zhang2023editguard}. LLM encryption, which uses word tokens and masks, faces difficulties in streaming contexts and struggles with varying textual styles, leading to potential false positives~\citep{yang2022tracing, kirchenbauer2023watermark}. Additionally, the removal of sentences can weaken the template’s presence, further challenging detection mechanisms~\citep{munyer2023deeptextmark}.

\Paragraph{Practical Implementation} Implementing these schemes also presents difficulties, particularly in balancing security with usability. Methods like sinusoidal signals and random isotropic unit vector perturbations aim to secure user data without compromising its utility but are often not robust against third-party overwriting~\citep{zhu2018hidden}. Furthermore, maintaining confidentiality in query results for models deployed as internal services is a challenge that attackers can exploit~\citep{hu2022protecting}. This balance between embedding strong security measures and preserving the practical utility of digital content remains a key issue.

In summary, while proactive schemes have introduced innovative means to secure digital content, they are limited by computational demands, vulnerabilities to attacks, and challenges in generalization and implementation.

\bibliographystyle{spbasic}
{\scriptsize\bibliography{sn-bibliography}}

\end{document}